\documentclass[10pt,letterpaper]{article}
\usepackage{algorithm}
\usepackage{algpseudocode}
\usepackage[top=0.85in,left=1.0in,footskip=0.75in]{geometry}

\usepackage{amsmath,amssymb}

\usepackage{changepage}
\usepackage[utf8x]{inputenc}
\usepackage{cite}
\usepackage{array}
\newlength\savedwidth
\newcommand\thickhline{\noalign{\global\savedwidth\arrayrulewidth\global\arrayrulewidth 2pt}%
\hline
\noalign{\global\arrayrulewidth\savedwidth}}
\usepackage{lastpage,fancyhdr,graphicx}
\newcommand{\N}{\mathbb{N}}
\newcommand{\R}{\mathbb{R}}
\DeclareMathOperator*{\argmin}{argmin}
\DeclareMathOperator*{\argmax}{argmax}
\DeclareMathOperator*{\mini}{minimize}

\begin{document}
\vspace*{0.2in}

\begin{flushleft}
{\Large
\textbf\newline{Laplacian mixture modeling for network analysis and unsupervised learning on graphs} 
}
\newline
\\
Daniel Korenblum\textsuperscript{1,*},
\\
\bigskip
\textbf{1} Research Division, Nanobio.Md, San Francisco, CA, United States of America
\bigskip

%
%





* d@nanobio.md

\end{flushleft}

\section*{Abstract}
\noindent Laplacian mixture models identify overlapping regions of influence in unlabeled graph and network data in a scalable and computationally efficient way, yielding useful low-dimensional representations.
By combining Laplacian eigenspace and finite mixture modeling methods, they provide probabilistic or fuzzy dimensionality reductions or domain decompositions for a variety of input data types, including mixture distributions, feature vectors, and graphs or networks.
Provable optimal recovery using the algorithm is analytically shown for a nontrivial class of cluster graphs.
Heuristic approximations for scalable high-performance implementations are described and empirically tested.
Connections to PageRank and community detection in network analysis demonstrate the wide applicability of this approach.
The origins of fuzzy spectral methods, beginning with generalized heat or diffusion equations in physics, are reviewed and summarized.
Comparisons to other dimensionality reduction and clustering methods for challenging unsupervised machine learning problems are also discussed.

\section*{Introduction}
Extracting meaningful knowledge from large and nonlinearly-connected data structures is of primary importance for efficiently utilizing data.
Big data problems (e.g. $> 1$ GB/s) often contain superpositions of multiple distinct processes, sources, or latent factors.
Estimating or inferring the component distributions or statistical factors is called the mixture problem.

Methods for solving mixture problems are known as mixture models \cite{ev96}, and in machine learning they are used to define Bayes classifiers  \cite{bishop}.
Mixture models are a widely applicable pattern recognition and dimensionality reduction approach for extracting meaningful content from large and complex datasets.
Only finite mixture models are described here, although countably or uncountably infinite numbers of mixture components are also possible \cite{jordan06}.
In terms of dimensionality reduction methods, Laplacian mixture models provide global and non-hierarchical analyses of massive datasets using scalable algorithms.
\subsection{Laplacian Eigenspace Methods}
Eigensystems of Laplacian matrices are widely used by spectral clustering methods \cite{azran}. 
Spectral clustering methods typically use the eigenvectors with small-magnitude eigenvalues as a basis for projecting data onto before applying some other clustering method on the projected item coordinates \cite{njw}.

\begin{figure}[!ht]
\centerline{\includegraphics[width=5.0in]{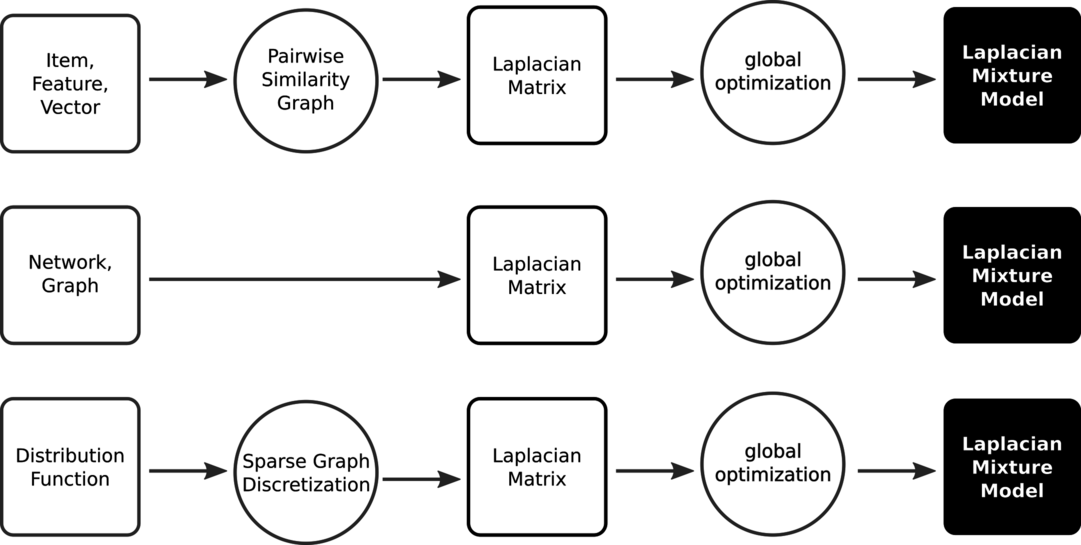}}
\caption{
{\bf Laplacian mixture modeling flow, gray squares show input datatypes and their mapping to Laplacian matrices (black square).}
Circles show processing steps, and the solid black square shows output model after globally optimizing the Laplacian eigenspace.}
\label{fig:0}
\end{figure}

In addition to graph/network data, Laplacian eigenspace methods can be applied to both discrete observation data and also continuous mixture density function data as shown in section \ref{sec:results}.
As Fig~\ref{fig:0} shows, feature vectors or item data are mapped to a graph via a distance or similarity measure \cite{nie2010efficient}, and mixture density data are mapped to a graph by
finite-difference approximations of the differential operator on a discrete grid or mesh.
Both feature vector data and continuous mixture density data are mapped to graph data as a preprocessing step prior to spectral graph cluster analysis.
For such applications, simple graphs are sufficient, meaning no self-loops or multiple edges of the same type are allowed.

When clustering data items, pairwise similarity or distance measures describe the regions of data space or subgraphs that represent closely related items.
In this context data are vectors, e.g. feature vectors in machine learning applications.
Laplacian eigenspace methods fall into the class of pairwise distance based clustering methods when data vectors are input.
It is the choice of this pairwise similarity or distance measure that is of utmost importance in creating accurate and useful results when generating Laplacian matrices from data items. One area of active research is in optimizing or learning the distance function based on some  training data \cite{gould14}.

Negative Laplacian matrices are also known as transition rate matrices or (infinitesimal) generators of continuous-time Markov chains (CTMCs), as first noted by \cite{dimitriev1945characteristic}.
Their exponentials are also referred to as heat kernels by analogy to the continuous heat equations that involve the continuous Laplace operator \cite{chung1998coverings,smola2003kernels,belkin2003laplacian}.
Heat kernels are also known as diffusion kernels, and have the same eigenvectors as Laplacians for discrete state spaces, or eigenfunctions for continuous state spaces \cite{kondor2002diffusion,bett2015diffusion}.

For many practical purposes, assuming that the chains are irreducible, meaning there is a path connecting every pair of states or nodes in the corresponding graph, does not lose any generality.
Strongly connected graphs correspond to irreducible chains, and chains can be broken into subchains and analyzed independently.
Finite-state CTMCs contain embedded discrete time Markov chains with related stochastic or Markov transition probability matrices with related properties to Laplacian matrices.
Given the holding times for each state, the stochastic matrix of the embedded chain is equivalent to the corresponding Laplacian for the CTMC.

On simple strongly connected networks, these matrices share the same first eigenvector (with different eigenvalues), called the Perron-Frobenius (PF) eigenvector or stationary probability distribution of the chain.
For non-symmetric Laplacians, the PF eigenvector can be either a right/column or a left/row eigenvector, depending on the matrix indexing convention used (right/column eigenvectors are used here).
In many cases of interest the Laplacians are symmetric, making this distinction irrelevant.
According to the Perron-Frobenius theorem, the PF eigenvector is always nonnegative and can be interpreted as a probability distribution.

Laplacian matrices both define distributions by their PF eigenvectors, and can also can be defined by distributions by constructing a matrix with a matching PF eigenvector.
This equivalence between distributions and Laplacian matrices provides a natural and useful bridge between probability distributions and Laplacian eigenspaces.
The duality between Laplacian matrices and probability distributions can be used for the purposes of statistical analyses and unsupervised machine learning.
Their spectral decompositions provide data-dependent bases for describing patterns that represent global, nonhierarchical structures in the underlying graph.

Laplacian mixture models are one way of probabilistically solving the multiple Laplacian eigenvector problem, as section \ref{sec:methods} describes.
They generate probabilistic mixture models directly from Laplacian eigenspaces by optimally combining other eigenvectors with the Perron-Frobenius eigenvector.
\subsection{Mixture Models}
Distinct component processes generate superpositions of overlapping component distributions when observed in aggregate, creating a mixture distribution.
The mixture problem is not easy to generally solve in part because it is so open-ended and difficult to objectively define in real-world contexts.
In 1894, Karl Pearson stated that \emph{the analytical difficulties, even for the case $n=2$ are so considerable, that it may be questioned whether the general theory could ever be applied in practice to any numerical case} \cite{pearson}.
Current unmixing or separation algorithms still cannot predict the number of components directly from observations of the mixture without additional information, or else they are parametric approaches that restrict components to fixed functional forms which are often unrealistic assumptions, especially in high dimensional spaces \cite{jordan06}.

Methods for separating the components of nonnegative mixtures of the form
\begin{equation}
f(x) = \sum_{k = 1}^{m} a_k f_k(x)\label{eq:fmm}
\end{equation}
are called mixture models, where $m \in \N$ is the number of mixture components and with $x \in \Omega$ an element of an index set e.g. $\Omega \subseteq \R^n$ in the continuous variable case or $\Omega \subseteq \mathbb{N}$ for the discrete case.
All of the results presented here for the continuous variable cases carry over to the discrete cases by replacing integrals $\displaystyle \int_{x \in \Omega} \bullet\ dx$ with summations $\displaystyle \sum_{x \in \Omega} \bullet $ for numerical accessibility.

Since all continuous problems must be discretized for numerical applications, the focus is on discrete variables $x$ with continuous problems saved for the appendix.
For all practical problems, it is safe to assume $f(x)$ is normalized as a probability distribution without loss of generality.

The $f_k(x)$ are known as the mixture components or component probability distributions of each independent process or signal source, and are also assumed to be normalized without loss of generality.
The $a_k \in [0, 1]$ are the weights or mixing proportions summing to one and forming a discrete probability distribution $P(k) = a_k$, which is known as the class prior distribution in the context of probabilistic or Bayes classifiers \cite{bishop}.

Finite mixture models can be used to define probabilistic classifiers and vice versa.
From exact knowledge of $\{f, f_1, \dots, f_{m}, a_1, \dots, a_{m}\},$ the posterior conditional distribution of an optimal Bayes classifier for any observed $y \in \Omega$ can be expressed as
\begin{align}
P(k\mid y) &= \frac{P(y\mid k) P(k)}{P(y)} \label{eq:pclass}\\
&= \frac{a_k f_k(y)}{f(y)},
\end{align}
forming a partition of unity over the space of observations or data \cite{fasshauer2007meshfree}.
The component distributions $f_k(x)$ can be understood as the class conditional distributions $P(x\mid k)$ and $f(x)$ as the evidence $P(x)$ in the context of  Bayes classifiers and supervised machine learning.
As probabilistic/Bayes classifiers $P(k\mid x)=\frac{a_k f_k(y)}{f(y)}$ form partitions of unity \eqref{eq:pclass} from finite mixture models \eqref{eq:fmm}, so do finite mixture models form partitions of unity as well.
In other words, partitions of unity can be defined first, without any reference to finite mixture models \eqref{eq:fmm}, as
\begin{equation}
p_k(x) \equiv \frac{a^\prime_k f^\prime_k(x)}{\sum_k a_k f_k^\prime(x)} \label{eq:poudef}
\end{equation}
subject to the constraints
\begin{equation}
\left\{
\begin{array}{cccc}
\displaystyle \sum_{k = 1}^m a^\prime_k &=& 1 &\\
\displaystyle \int_\Omega f^\prime_k(x) dx &=& 1 & k = 1, \dots, m.
\end{array}\label{eq:poudefcon}
\right.
\end{equation}
To connect mixture models with partitions of unity, the mixture components $\{f_k(x)\}_{k = 1}^m$ and weights $\{a_k\}_{k = 1}^m$ from mixture models \eqref{eq:fmm} can be related to \eqref{eq:poudef} for $k=1, \dots, m$ according to
\begin{equation}
\left\{
{\setlength{\extrarowheight}{8pt}
    \begin{array}{ccc}
        a_k(x) &=& \displaystyle\int_\Omega p_k(x) f(x) dx\\
        f_k(x) &=& a_k^{-1} p_k(x) f(x)\\
    \end{array}
}
\right.
    \label{eq:waf}
\end{equation}
with $\displaystyle \sum_{x \in \Omega} p_k(x) f(x)$ for discrete $\Omega$ cases.
This explicitly shows the formal equivalence of partitions of unity \eqref{eq:poudef} as discriminative versions of finite mixture models \eqref{eq:fmm}.
In this case, the partition of unity $\{p_k(x)\}_{k = 1}^m$ can be interpreted as $\{P(k\mid x)\}_{k = 1}^m$, the mixture component conditional probabilities.	

Partitions of unity \eqref{eq:poudef} do not explicitly involve $f(x)$ the mixture distribution, or require it as an input.
In other words, the component conditional probabilities $\{p_k(x) = P(k\mid x)\}_{k = 1}^m$ can still be computed even if knowledge of $f(x)$ the underlying mixture distribution is not available or is not required.
This makes the partition of unity form of mixture models \eqref{eq:poudef} very useful in practice, since they apply even in cases where estimating $f(x)$ is not relevant or available such as cluster analysis, graph partitioning, and domain decomposition.
Therefore, mixture models can be considered as a special case of partitions of unity applied to the separation of mixture probability distributions.

In applications such as cluster analysis or graph partitioning, unlabeled data are automatically assigned to various groups known as clusters to distinguish them.
Clusters can be understood as unsupervised analogs of classes from supervised machine learning.
In cluster analysis, computing a partitions of unity of the form \eqref{eq:poudef} provides a probabilistic or fuzzy clustering or partition.
The components of the partition of unity $\{p_k(x)\}_{k = 1}^m$ can be interpreted as conditional probabilities over the clusters or partitions, analogous to the class conditional distributions \eqref{eq:pclass} of Bayes classifiers.

For such problems the mixture components $\{f_k(x)\}$ are not relevant, and only the cluster conditional probabilities $\{p_k(x)\}$ are needed.
Formally, they can be viewed as mixture models via \eqref{eq:poudef} where $f(x) = constant$ the uniform distribution over $\Omega$ the domain.
Soft clusterings are most useful when insights into the global structures of data spaces or networks are of interest.
Hard clustering algorithms, such as $k$-means, do not provide any information about the global, nonhierarchical relationships between items or nodes in a graph.
Rather than being dichotomous as implied by their names, soft and hard clustering approaches are complementary, and may be used together in one analysis to answer different questions about the same dataset.
\section*{Materials and methods}\label{sec:methods}
Finite mixture models of the form \eqref{eq:fmm} are easy to understand and interpret due to their probabilistic definition.
This motivates hybridizing finite mixture models and Laplacian eigenspace methods, as described in this section.
\subsection{Notation}
Let $V \equiv \{ 1, \dots, N\}$ be the set of ordered vertices of a simple strongly connected graph $G$ with weight function $w: V \times V \to [0, 1]$, and $w(i, j)=0$ if no edge between vertices $i$ and $j$ exists.
Let $A\in\R^{N\times N}$ be the weighted adjacency matrix of $G$ with $a_{ij} = w(i, j)$, and let $d \equiv e^T A$ and $D \equiv \mathrm{diag}\ (d_1 \cdots d_N) \in \R^{N \times N}$ be the corresponding diagonal weighted degree matrix.
$e \in \R^N$ represents the column vector of all ones.
The weighted Laplacian $\Delta \in \R^{N \times N}$ of graph $G$ for some fixed ordering on the vertices $V$ is given by
\begin{equation}
\Delta(u, v) \equiv
\left\{
  \begin{array}{ll}
    d_v, & \mathrm{if}\ u = v; \\
    -w(u, v), & \mathrm{if}\ u\ \mathrm{and}\ v\ \mathrm{are}\ \mathrm{adjacent}; \\
    0, & \mathrm{otherwise},
  \end{array}
\right.
\end{equation}
where $d_v$ is the weighted vertex degree of the $v$-th vertex, and $u = 1, \dots, N$, $v = 1, \dots, N$.
In matrix notation this becomes $\Delta = D - A$.
Using this definition, the Perron-Frobenius vector corresponds to the right column eigenvector of $\Delta$ with eigenvalue zero.

Let $\phi_i(x), i=0, \dots, N-1$ denote the right eigenvectors of $N \times N$ Laplacian matrices $\Delta$ of simple strongly connected weighted or unweighted graphs, and similarity for their continuous eigenfunction analogs where $N = \infty$.
Since the Laplacian matrices $\Delta$ are assumed to be normal, their right eigenvectors form a complete orthonormal basis over $\Omega$, the discrete or continuous domain.
Assume that these eigenvectors are ordered according to ascending eigenvalue magnitude.
The Laplacian mixture modeling algorithm estimates the true finite mixture distribution components $f_k(x)$, $k = 1, \dots, m$ from \eqref{eq:fmm} directly from $\{\phi_0, \dots, \phi_{m - 1}\}$ the first $m$ right eigenvectors of normal Laplacian matrices.
For now, assume that $m$ is known or given as a constant Laplacian mixture models are determined directly from the, with $m \ll N$ in many dimensionality reduction problems \cite{nie2010flexible,7976386}.
Section \ref{sec:results} describes methods for estimating $m$ using existing model selection techniques in cases where it is not known.

Although the algorithm is defined for continuous domains, the scope is limited to discrete problems here, to focus on the most practical situations.
The discrete input is the set of eigenvectors $1, \dots, m$ of $\Delta$, the weighted graph Laplacian.
Unweighted graphs occur for binary weights $w: V \times V \to \{0, 1\}$ that are all zeros or ones, which can be considered a special case of continuous weights on the unit interval.
The assumption that $w$ has a maximum value of 1 can be made without loss of generality.

Three types of input data and their corresponding analysis problems are considered:
\begin{enumerate}
  \item Graph or network data are converted to Laplacian matrices via their weighted adjacency matrices, and the problem is to infer the optimal fuzzy assignments or soft partitioning for community detection and centrality scoring.
  \item Feature vector data, which are converted into Laplacian form using pairwise similarity or distance measures, in which case the problem is to estimate $\{p_k(x)\}_{k = 1}^m$ the conditional mixture probability estimates.
  \item Sampled values of a mixture density function $f(x)$, for which the resulting Laplacian is designed so that $f(x) \equiv \phi_0(x)$, the Perron-Frobenius or first Laplacian eigenvector, and the problem is to estimate $\{f_k(x)\}_{k = 1}^m$ the mixture components.
\end{enumerate}
The vector of fuzzy spectral estimates $\hat p$ for the true values $p_k(x)$ from \eqref{eq:poudef} can be defined in terms of a nonlinear optimization problem for $M \in \mathrm{GL}(m) \subset \R^{m \times m}$, the square invertible $m \times m$ matrix of expansion coefficients.

Minimizing the error or nondeterminicity of the model from a Bayes classifier standpoint \eqref{eq:pclass} serves as an objective for determining the optimal matrix $M^{*}$ having the least total overlap of the macrostate boundaries allowable by the subspace spanned by the selected right eigenvectors.
The sum of the expected values of the squares of the conditional probabilities
\begin{equation}
\sum_{k = 1}^m \left\langle p^2_k(x) \right\rangle_{\phi_0}
\end{equation}
equals one if and only if they are perfectly binary or deterministic.
Therefore the deviation of the expected squares of the conditional probabilities from unity $1 - \sum_{k = 1}^m \left\langle p^2_k(x) \right\rangle_{\phi_0}$ serves as a measure of the squared error i.e. fuzziness, overlap, or nondeterminicity.
\subsection{Definition}\label{sec:lemmdef}
Let the loss function $0 < L < 1$ represent this expected error or nondeterminicity i.e. areas of fuzziness or overlap where the conditional probabilities are non-binary,
\begin{equation}
L    \equiv 1 - \sum_{k = 1}^m \left\langle \hat p^2_k(x) \right\rangle_{\phi_0} \label{eq:loss}
\end{equation}
which is equivalent to the probabilistic classifier's expected error using a quadratic loss function \cite{bishop}.
($\int_\Omega \bullet dx$ can be replaced by $\sum_{x \in \Omega} \bullet$  in discrete cases.)
This objective function definition connects Laplacian eigenspace methods and probabilistic/Bayes classifiers \eqref{eq:pclass} derived from finite mixture models.

The minimum expected error condition
\begin{equation}
M^{*} \equiv \argmin_{M}L(M) \label{eq:mopt}
\end{equation}
subject to the partition of unity constraints
\begin{align}
\left\{
{\setlength{\extrarowheight}{6pt}
\begin{array}{ccccc}
\hat p_k(x) &\geqslant& 0,&\forall x \in \Omega,&\ k = 1, \dots, m\\
\displaystyle \sum_k \hat p_k(x) &=& 1,&\forall x \in \Omega&
\end{array}
}
\right.\label{eq:pou}
\end{align}
selects maximally crisp (non-overlapping) or minimally fuzzy (overlapping) decision boundaries among classifiers formed by the span of the selected right eigenvectors.
This objective function attempts to minimize the expected overlap or model error between the component distributions of the mixture distribution $f(x) \equiv \phi_0(x)$ defined by the PF eigenvector of the input Laplacian.
Minimizing the loss $L(M)$ occurs over the matrix of expansion coefficients for the eigenspace spanned by the selected right eigenvectors $\{\phi_i\}_{i = 0}^{m - 1}$ which serve as an orthogonal basis.
The eigenbasis provides linear inequality constraints defining a feasible region in terms of a convex hull over $M \in \mathrm{GL}(m)$ the expansion coefficient parameter space.

Therefore
\begin{equation}
\hat f_k(x; M^{*}) \equiv \hat a_k^{-1}(M^{*}) \hat p_k(x; M^{*}) f(x) \label{eq:fw}
\end{equation}
provides the following definition of a \emph{Laplacian mixture model} via \eqref{eq:waf}:
\begin{equation}
f(x) = \sum_{k = 1}^m \hat a_k(M^{*}) \hat f_k(x; M^{*}),\label{eq:mmm}
\end{equation}
where $M^{*}$ solves \eqref{eq:mopt} and \eqref{eq:pou}, a linear-constrained concave quadratic optimization problem.
\subsection{Global Optimization}\label{sec:gopt}
In terms of probabilistic classifiers, the loss function $L(M)$ represents the mean squared error of the probabilistic classifier derived from the mixture model specified by each value of $M$.
By minimizing $L(M)$, Laplacian mixture models provide spectrally regularized minimum mean squared error separations for a variety of input data types, as shown in Fig~\ref{fig:0}.

For any viable value for $m$ the number of mixture components, the column vector valued function
\begin{equation}
\hat p \equiv \left(\begin{array}{ccc}\hat p_1(x; M) & \cdots & \hat p_m(x; M) \end{array}\right)^T
\end{equation}
is numerically optimized over $M$ to compute the Laplacian mixture model.
The optimal $\hat p$ is determined via global optimization of $L(M)$ over the set of coordinate transformation matrices $M$ satisfying the partition of unity constraints $\eqref{eq:pou}$ to compute $M^*$ the globally-optimized model parameters.

In matrix notation, any feasible $\hat p$ can be written in terms of $M$ and the column vector of basis functions
\begin{equation}
\omega \equiv \left(
  \begin{array}{c}
    \frac{\phi_0(x)}{\sqrt{\phi_0(x)}}\\ \\
    \frac{\phi_1(x)}{\sqrt{\phi_0(x)}}\\
    \vdots\\
    \frac{\phi_{m - 1}(x)}{\sqrt{\phi_0(x)}}
  \end{array}
\right)\label{eq:omega}
\end{equation}
as
\begin{equation}
\hat p = M \omega
\end{equation}
subject to
\begin{equation}
  \left\{
{\setlength{\extrarowheight}{4pt}
  \begin{array}{cl}
    M^T e &= e_1\\
    M \omega (x) &\geqslant 0\ \forall x,
  \end{array}
}
\right.
\end{equation}
where $e = \left( \begin{array}{cccc} 1 & 1 & \cdots & 1\end{array}\right)^T$ and $e_1 = \left( \begin{array}{cccc}1 & 0 & \cdots & 0\end{array}\right)^T$ are $m \times 1$ column vectors.

Now the objective function $L(M)$ can be expressed in terms of $\omega$ and $f(x) \equiv \phi_0(x)$ as
\begin{align}
L &= 1 - \sum_{k = 1}^m \left\langle \hat p_k^2(x) \right\rangle_{\phi_0}\\
&= 1 - \left\langle \omega^T M^T M \omega \right\rangle_{\phi_0}\\
&= 1 - \sum_{i, j = 0}^{m - 1} \left(M^T M\right)_{ij} \underbrace{\int_\Omega \phi_i(x) \phi_j(x) dx}_{\delta_{ij}} \\
&= 1 - \mathrm{tr}\ M^T M,
\end{align}
a concave quadratic function where $\mathrm{tr}\ M^T M$  is equal to $\| M \|_F^2$, the squared Frobenius norm of $M$ the eigenbasis transformation matrix.

Weighted graph Laplacians may not always be normal matrices, in which case the eigenvectors may not be orthogonal.
In many cases, the Laplacian can be symmetrized prior to running the Laplacian mixture modeling algorithm.
Directed weighted graphs with combinatorially symmetric adjacency matrices correspond to finite ergodic (i.e. irreducible and aperiodic) Markov chains satisfying detailed balance.
Combinatorially symmetric graph Laplacians can be symmetrized by conjugation with diagonal matrices, analogous to a change of variables in the corresponding heat equation \cite{kramer1959symmetrizable,parter1962symmetrization}.
Reversing the change of variables after computing the Laplacian mixture model allows the same form as above to be used in this more general context.
This type of symmetrized Laplacian has the same eigenvalues as the original unnormalized graph Laplacian, unlike the standard symmetric normalized graph Laplacian \cite{chung1997spectral}.

These results allow the linearly constrained global optimization problem for $M^{*}$ to be stated as
\begin{equation}
\begin{array}{cc}
\displaystyle \mini_M & 1 - \| M \|_F^2 \\
& \\
\mathrm{subject\ to} & \\
& {\setlength{\extrarowheight}{2pt} \begin{array}{cccc} M^T e &=& e_1& \\ \left(M \phi\right)(x) &\geqslant& 0& \forall x \\ M &\in& \mathrm{GL}(m). &\end{array} }
\end{array}
\label{eq:gopt}
\end{equation}
The linear inequality constraints $M \phi \geqslant 0$ encode all of the problem-dependent data from the input Laplacian via
\begin{equation}
\phi = \left( \begin{array}{cccc} \phi_0(x) & \phi_1(x) & \cdots & \phi_{m - 1}(x) \end{array} \right)^T.
\end{equation}

This linearly constrained concave quadratic problem is an archetypal example of an NP-hard problem because of the combinatorial number of vertex solutions defined by the convex polytope formed by the constraints.
Statistically useful solutions are not guaranteed to exist and even then they do, approximating the solution to \ref{eq:gopt} requires specialized numerical algorithms.
Simple yet nontrivial problems where this optimization problem can be solved analytically, yielding optimal models, are shown next.
This provides evidence for the usefulness of Laplacian mixture models, and highlights some of their spectral graph theoretic properties.
\subsection{Interpolating cluster graphs}
In this section, Laplacian mixture models are shown to perform optimally on a type of graph that can be called interpolating cluster graphs.
Cluster graphs are unions of complete graphs, and have block diagonal adjacency matrices, corresponding to disjoint subgraphs.
Complete graphs have adjacency matrices whose off-diagonal elements are all equal to one, corresponding to fully interconnected vertices.
Cluster and complete graphs are both well studied and understood from the perspective of spectral graph theory \cite{shamir2004cluster}.

Interpolating them allows spectral graph theoretic results for cluster and complete unweighted graphs to be extended to cluster-weighted complete graphs.
Suppose that $K \in \N_{\geqslant 2}$ is the number of blocks, $N_k \in N_{\geqslant 2}$, $k = 1, \dots, K$, are the vertex counts for each cluster, and $N \equiv \sum_k N_k$ is the total vertex count.
Let the family of interpolating cluster graphs $B \in \R^{N \times N}$ be defined by their adjacency matrix form:
\begin{align}
B &\equiv A_{cluster} + \left( 1 - \varepsilon \right) \left[ e_N e_N^T - A_{cluster} \right]\label{eq:interp}\\
A_{cluster} &\equiv \bigoplus_{k = 1}^K \left[ e_{N_k} e_{N_k}^T - I_{N_k} \right].
\end{align}
The matrix $I_n \in \R^{n \times n}$ denotes the $n$-by-$n$ identity matrix and $e_n e_n^T \in \R^{n \times n}$, $n \in \N_{\geqslant 2}$, denotes the outer product of the real vector of all ones with itself, i.e. the rank-one $n$-by-$n$ matrix of all ones.

Although the term ``interpolating graph'' was introduced here for the specific case of interpolating between cluster graphs and complete graphs, the concept can be defined for any two arbitrary graphs with the same nodes.
In general, given two $N \times N$ graph adjacency matrices $A_0$ and $A_1$, an interpolating graph adjacency matrix $A_{interp} = ( 1 - \varepsilon ) A_0 + \varepsilon A_1$ can be defined as their convex sum.
Using this form, interpolating cluster graphs \ref{eq:interp} can be written $B = (1 - \varepsilon) A_{complete} + \varepsilon A_{cluster}$, where $A_{complete}$ and $A_{cluster}$ are the respective $N \times N$ adjacency matrices for a complete graph and a cluster graph.
This shows that interpolating cluster graphs form a connection between complete and cluster graphs.

Therefore, interpolating graphs can be seen as a specific type of graph perturbation.
Graph perturbations such as those involving mass-spring networks have been studied in \cite{maas1987perturbation,rowlinson1990more,guo2007laplacian}.
Studying the properties interpolating graphs in more depth may be of theoretical interest, but that is not the focus of this article.

The adjacency matrices of interpolating cluster graphs interpolate between the adjacency matrices of cluster graphs and the adjacency matrices of complete graphs via $\varepsilon \in [0, 1]$, a separation parameter.
Interpolating cluster graphs are complete graphs when the separation parameter $\varepsilon$ equals zero, and they are cluster graphs for $\varepsilon=1$.
In between, for $0 < \varepsilon < 1$, interpolating cluster graphs provide models of highly similar of vertices representing homogeneous network communities.

When $\varepsilon = 1$, $B = A$, a block diagonal adjacency matrix corresponding to a cluster graph that is a union of $K$ smaller complete $N_k$-vertex graphs.
Results from spectral graph theory show that for cluster graphs, the zero eigenvalue has multiplicity $K$ and that the connected components of the graph can be directly identified from the position of their nonzero elements.
All of the information about the structure of cluster graphs is contained in the $K$-dimensional Laplacian eigenspace associated with the zero eigenvalue.

With $\varepsilon = 0$, $B = e_N e_N^T - I_N$, the adjacency matrix of a complete $N$-vertex graph with uniform edge weights.
Spectral graph theory shows that the multiplicity of the 2nd eigenvalue of a complete graph is $N - 1$, and its value is $N$.

The 2nd Laplacian eigenvalue, called the Fiedler value, measures the connectivity of the graph and can be used to find optimal partitions \cite{fiedler1973algebraic}.
When the Fiedler value has multiplicity one, its corresponding eigenvector is known as the Fiedler vector.
The eigendecomposition of interpolating cluster graphs was found to have a simple analytically solvable form.

Interpolating cluster graphs have a 2nd Laplacian eigenvalue with multiplicity $K-1$, so in some sense, they have $K - 1$ Fiedler vectors.
This makes using the Fiedler vector to identify the corresponding cluster partitions a more challenging problem.
In this section, Laplacian mixture models are shown to optimally solve it.

For any $0 \leqslant \varepsilon < 1$, the graph is connected, the zero eigenvalue of $B$ has multiplicity one, and the eigenvector with eigenvalue zero is the constant vector.
The one-dimensional Laplacian eigenspace associated with the zero eigenvalue no longer contains any information about the cluster structure of the graph.
In addition, the 2nd eigenvalue equals $N ( 1 - \varepsilon )$ with multiplicity $K-1$, and the eigenvectors associated with this eigenvalue are not uniquely determined.
It is no longer obvious how to infer the cluster assignments of each vertex directly from the eigenvectors associated with the first $K$ eigenvalues.

For interpolating cluster graphs, the Laplacian mixture modeling optimization problem \eqref{eq:gopt} can be analytically solved, as shown below.
Laplacian mixture models recover the structure of interpolating cluster graphs exactly, converting the first $K$ eigenvectors into binary conditional probabilities for belonging to each cluster.
Although the eigenvectors associated with the 2nd eigenvalue are not uniquely determined, expressions for them can be defined to encode the transition rates of conserved quantities between blocks.

Choosing the first cluster as a reference for concreteness, the analytic expressions for these $K - 1$ eigenvectors, $\phi_{ij}$, where $i=1, \dots, N$, and $j = 0, \dots, K - 1$, are given by:
\begin{equation}
\phi_{ij}=\left\{
  \begin{array}{cl}
    1, & \hbox{if vertex $i$ is in cluster 1;} \\
    -\frac{N_1}{N_k}, & \hbox{if vertices $i$ and $j$ are in the same cluster;} \\
    0 & \hbox{otherwise.}
  \end{array}
\right.
\end{equation}
The corresponding 2nd eigenvalue with multiplicity $K - 1$ equals $N ( 1 - \varepsilon )$.
These expressions can be easily verified e.g. by direct substitution or using symbolic manipulation tools for specific choices of $K$.

Although the eigenvectors listed above are not mutually orthogonal, they are orthogonal to the constant vector, since they sum to zero.
They are also linearly independent, and can therefore be used to prove the optimality of the Laplacian mixture model for this graph.
Putting
\begin{align}
p_1 &= \frac{1}{N} \sum_{i = 1}^K N_i \phi_{i - 1}\\
p_j &= \frac{N_j}{N_1} ( p_1 - \phi_{j - 1} ),\ \forall j=0, \dots, K - 1,
\end{align}
solves the Laplacian mixture modeling problem explicitly, in closed-form for any finite $K < \infty$ number of blocks.

These binary conditional probabilities form an indicator matrix $p \in \R^{N \times m}$ for the cluster assignments of each vertex.
I.e., $p_{ij} = 1$ if vertex $i$ is in cluster $j$ and 0 otherwise, providing an exact solution to the cluster weighted complete graph analysis problem.
This corresponds to an ideal solution with no fuzziness in the assignments of each vertex to a single cluster.
The objective function value $L$ from \eqref{eq:loss} reaches its lower bound of zero for such binary mixture component conditional probabilities.
Laplacian mixture models therefore have provable optimality properties on cluster interpolating graph types.

For this analytically solvable case, Laplacian mixture models exactly recover the block diagonal structure of $B\left(\varepsilon; K, \{N_k\}_{k = 1}^K\right)$ the interpolating cluster graphs, for all cluster numbers $K$, cluster sizes $\{N_k\}_{k = 1}^K$, and separation parameters $\varepsilon \in (0, 1)$.
For any fixed $K$, This can be easily verified directly or via symbolic manipulation tools.
Rigorous proofs, using mathematical induction, are beyond the scope of this article.
Having solutions to Laplacian mixture modeling problems for any $B\left(\varepsilon; K, \{N_k\}_{k = 1}^K\right)$ on arbitrarily large graphs and cluster numbers allows validation of numerical solvers.

In the limit of vanishing separation parameter $\varepsilon$, the graph approaches a complete graph without any cluster structure.
When $\varepsilon$ becomes arbitrarily close to zero, the 2nd eigenvalue becomes arbitrarily close to the higher eigenvalues, and the gap between the 2nd and higher eigenvalues becomes arbitrarily small.
Laplacian mixture models remain exact for this class of graphs regardless of the magnitude of the spectral gap, i.e. even when there are no sparse cuts.
These analytic results show that there are cases where Laplacian spectral gap presence is sufficient but not necessary for the first $K$ eigenvectors to be useful in identifying graph structure.

Theoretical worst-case analysis suggests that exact solutions to the NP-hard global optimization problem involved are not possible.
Nevertheless, useful approximate solutions have been developed for density function estimate, feature/vector, and graph data during the course of numerical testing.
Numerical examples are presented next to demonstrate the flexibility, versatility, and other features of Laplacian mixture modeling approaches.
\section*{Results and Discussion}\label{sec:results}
The NP-hard nature of the linearly-constrained concave quadratic global optimization problem makes exhaustively computing all possible models is impractical, due to the geometric growth with increasing model dimension in the number of vertices in the convex hull defined by the linear inequality-constraints.
Empirical testing suggests that for most applications, this theoretical obstacle can be avoided in practice, and accurate solutions sampling strategies can be scalably implemented for approximating the global optimizer on large datasets.
In part, this is due to the detailed information contained in each of the factors or mixture components identified due to their fuzzy/overlapping/probabilistic form.

Laplacian mixture model components identify large-scale regions of the graph in such a way that each region covers the entire graph.
Each mixture component, factor, or dimension provides information about the entire graph whose weighting favors the region it corresponds to.
As demonstrated in this section, it is possible to efficiently compute a reasonable range of components or dimensions even on relatively large graphs, making the NP-hard aspects of the global optimization problem circumventable.

Optimizing the sampling strategy for approximately solving the linearly-constrained concave quadratic global optimization problem is an open problem not discussed here.
These parameters are used as inputs for the modified Frank-Wolfe heuristic, which involves solving a set of linear programming problems.
The details of the modified Frank-Wolfe heuristic are described in more detail in \cite{pard93}.
Since they are not data-dependent, and for the sake of consistency when comparing different runs, the global optimization search parameters were precomputed and stored in a lookup table.
By using a lookup table, computing these parameters does not contribute any significant overhead to the optimization runtimes.
The current strategy is to gain confidence in the reliability of the approximant by locating the same one multiple times using different search parameters by slightly oversampling the solution space.

Because of numerical limits of finite-precision arithmetic, it is impossible to perfectly enforce the constraint that the transformation matrix $M^*$ be invertible.
One way of dealing with this issue is to apply an even stricter constraint that often makes sense for data analysis applications.
This constraint requires that all factors contain at least one item whose probability for that factor is larger than all other factors.

This means that hard-thresholding the conditional factor probabilities must not create any degenerate clusters with zero items assigned to them.
It's a reasonable criterion for applications where each factor represents an independent observable signal generating process.
Therefore, the optimal solution is chosen by minimizing the loss function $L(M)$ over all non-degenerate models.
Many of the computed solutions are degenerate, an aspect of the NP-hard optimization problem that is partially circumvented in practical situations using heuristics to accurately approximate the solution.

Algorithm \ref{alg:1} lists the pseudocode describing the actual operations performed for all of the numerical results presented here.
\begin{algorithm}[!ht]
	\caption{Compute $p$ for $N\times m$ Laplacian eigenvector matrix $\phi$}
	\begin{algorithmic}[1]
		\Procedure{LaplacianMixtureModel}{$\phi, S$}
   			\State Initialize scalar $L_{min} \leftarrow \infty$
   			\State Initialize $m \times m$ array $M^*$
   			\For {$s \in S$} \Comment{Run the modified Frank-Wolfe heuristic}
       			\State $M_s \leftarrow \displaystyle \argmin_{M} \mathrm{LP}(M; \phi, s)$
       			\State $p^T \leftarrow M_s \phi^T$
       			\State Initialize $1 \times m$ array $c \leftarrow 0$
       			\For {$i = 1, \dots, N$} \Comment{Run the degeneracy check}
       			  \State $j \leftarrow \displaystyle \argmax_{k = 1, \dots, m} \left( p^T_i \right)_k$
       			  \State $c_j \leftarrow c_j + 1$
       			\EndFor
       			\If{$\displaystyle \min_{k = 1, \dots, m} c_k > 0$}
      	    		  \State $L(M_s) \leftarrow 1 - \| M_s \|_F^2$
       			\Else
     			      \State $L(M_s) \leftarrow \infty$
       			\EndIf
       			\If{$L(M_s) < L_{min}$}
       			    \State $L_{min} \leftarrow L(M_s)$
       			    \State $M^* \leftarrow M_s$
       			\EndIf
   			\EndFor
   			\State $p^T \leftarrow M^* \phi^T$
   			\State Return $\{ p, M^* \}$
		\EndProcedure
	\end{algorithmic}\label{alg:1}
\end{algorithm}
The details of the modified Frank-Wolfe heuristic are described in \cite{pard93}.
In this section, several numerical examples are presented to illustrate some Laplacian mixture model sampling strategies and demonstrate their efficacy on both synthetic and real-world problems.
The results show high levels of performance across a range of problem types without encountering problems such as numerical instability or sensitivity to small variations of tunable parameters.
All model computation run-times were less than 24 hours using {\sc Matlab} on a dual 2.40 GHz {\sc Intel Xeon E5-2640} CPU running the 64-bit {\sc Windows 10} operating system with 128GB of RAM.

Data are cleared from memory after the sparse similarity matrix is computed to conserve resources.
Small-magnitude eigenvector estimates for $\{\phi_i(x)\}_{i = 0}^{m_{max}}$ are computed using the Matlab function \texttt{eigs}, a sparse iterative solver.
The similarity matrix may be cleared from memory after computing the first $m$ eigenvectors, before the spectral mixture models are computed.
\subsection*{Noisy interpolating cluster graphs}
The analytic solution of the spectral mixture modeling problem for interpolating cluster graphs provides provable optimality results for these class of idealized noise-free graphs.
Performance evaluation in the presence of noise is also necessary in order for these theoretical results to be meaningful on real graph datasets, which contain noise.

It is not possible to test every possible combination of cluster number, cluster size distribution, noise level, and noise type.
Nevertheless, testing with a few varied noise levels and combinations is sufficient to provide evidence for the stability and robustness of the algorithm performance.

To demonstrate the algorithm's multiscale pattern recognition capabilities, five log-linearly spaced cluster sizes were chosen, ranging from 2 to 10,000 vertices.
Six varied noise levels corresponding to choices of the $\alpha$ uniform noise parameter were also tested.
Uniform noise was symmetrically added to noise-free interpolating cluster graph adjacency matrices $B(\varepsilon)$ according to
\begin{equation} \tilde b_{ij} = b_{ji} \equiv (1 - \alpha) b_{ij} + \alpha u \end{equation}
to generate noisy adjacency matrices $\tilde B(\varepsilon)$, where $u$ is a sample from the uniform distribution over the unit interval.
This form ensures that all elements of $\tilde B$ lie between zero and one.
(Matlab code implementing the generation of these noisy interpolating cluster graphs is available upon request.)

Sparsifying the graph prior to analysis by using a cutoff parameter on the nearest neighbors of each vertex is common in practice.
See \cite{von2007tutorial} for details on the intuition and justification for this sparsification parameter.
Five different values of the $k_{NN}$ nearest neighbor parameter and three different values of the $k_{min}$ outlier removal parameter were applied to validate the methodology subsequently used for real world data.
The entire test was repeated for $N_{out} = 0$ and $N_{out}=1$ choices of $N_{out}$, the outlier noise parameter.

\begin{figure}[!ht]
\centerline{\includegraphics[width=4.5in]{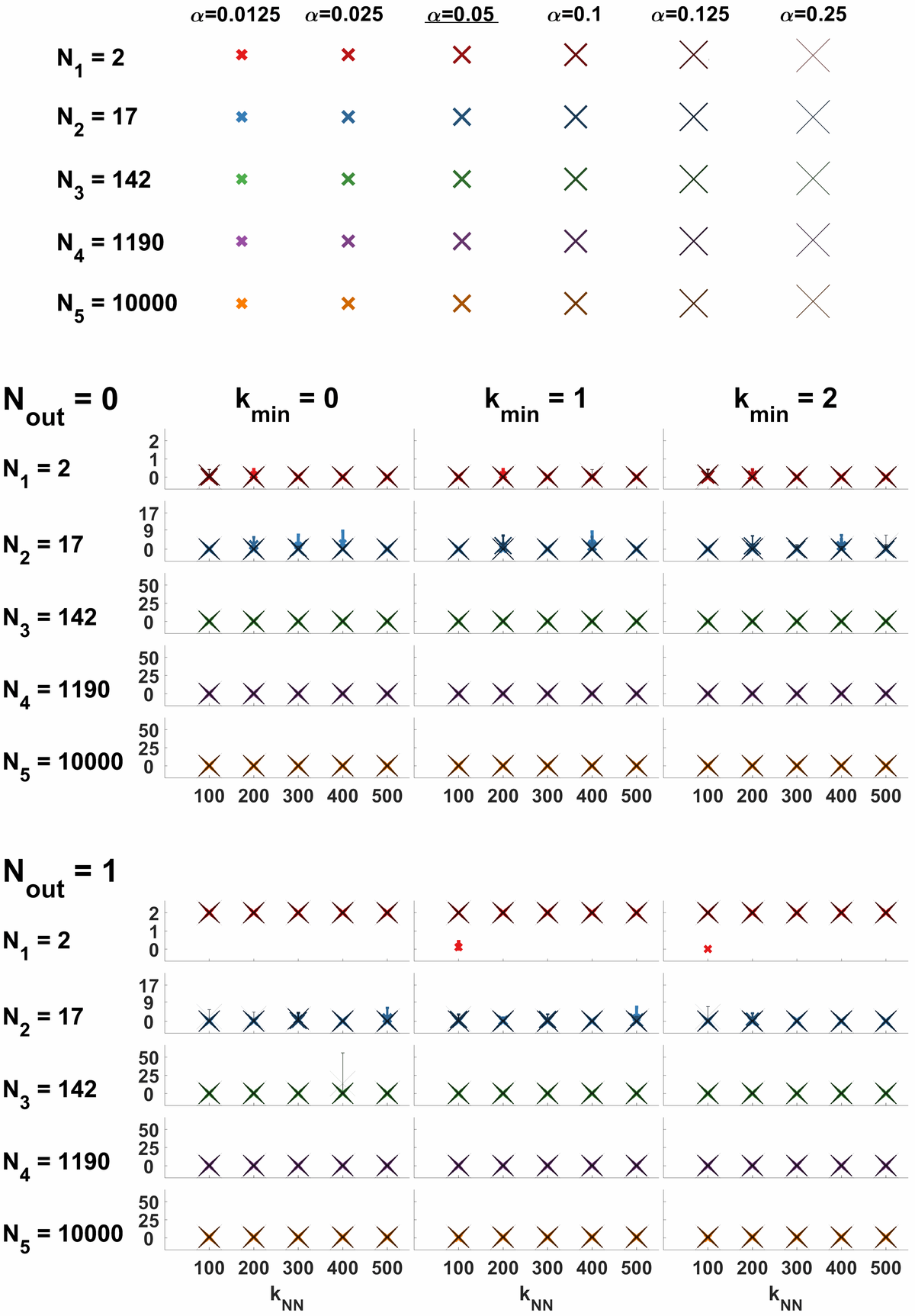}}
\caption{
{\bf Overlay plot matrix of number of errors vs. $k_{NN}$, representing a total of 1800 Laplacian mixture modeling algorithm runs.}
Noisy interpolating cluster graph matrices $\tilde{B}\left( \varepsilon = 1/2 \right)$ were randomly generated for each run.
Markers represent sample averages of labeling error numbers, after assigning each vertex to one of the 5 mixture components identified.
Marker sizes correspond to uniform noise level $\alpha$ as indicated by the marker key on top of the plot matrix.
Error bar lengths indicate one sample standard deviation.
Colors correspond to the different log-linearly spaced cluster sizes $(2, 17, 142, 1190, 10^4)$ as labeled.
Columns correspond to different values of $k_{min}$, the minimum strongly-connected neighbor number.
Outlier noise level indicated by $N_{out}$ parameter separating upper and lower sections of the plot matrix.
}
\label{fig:1}
\end{figure}

For each of the resulting 180 $\{\alpha, k_{NN}, k_{min}, N_{out}\}$ parameter combinations, ten randomized samples were drawn using the same pseudorandom seed to allow direct comparisons to be made.
Panels of Fig~\ref{fig:1} show the number of errors vs. $k_{NN}$ as errorbar overlay plots, representing a total of 1800 runs.
Markers represent average labeling errors after assigning each vertex to one of the 5 mixture components identified.
Error bar lengths indicate one sample standard deviation, and colors correspond to the different cluster sizes as labeled.

In the absence of outliers ($N_{out} = 0$), perfect accuracy was achieved for all noise levels on cluster sizes larger than 100.
For some choices of $k_{NN}$, for any choice of $k_{min}$, perfect accuracy was obtained for all cluster sizes over all samples, showing that setting $k_{min} > 0$ does not introduce algorithm errors.
In fact, perfect accuracy was achieved at $N_{out} = 1$ for all cluster sizes and noise levels using $k_{min} = 1$ for several values of $k_{NN}$.

For one outlier ($N_{out} = 1$), the accuracy of the algorithm on the smallest cluster size of 2 suffers significantly at all noise levels, as shown in the lower section of Fig~\ref{fig:1}.
In this case, setting $k_{min}=1$ significantly improved the accuracy of the smallest cluster size with 2 vertices and low noise levels (top panel, middle column), reducing the average errors count to 0.1.

Fig~\ref{fig:1} shows that although high levels of noise make small cluster sizes difficult to recover, overall, the best accuracy occurs with $k_{min} = 1$ and $k_{NN} = 100$.
The improvements in accuracy shown for $N_{out}=1$ using $k_{min}=1$ were also verified using higher values of $N_{out}$ (data not shown).
This indicates that the algorithm's optimal recovery property for cluster graphs is highly robust to noise and is not overly sensitive to the choice of $k_{NN}$ and $k_{min}$ parameters.
It also supports the methodology of using  $k_{min} > 0$ and $k_{NN} < N$ for real data, as described in subsequent sections.
\subsection*{Data Clustering/Fusion}
Only the eigenvectors with small magnitude eigenvalues are necessary to compute Laplacian mixture models, which are of lower dimensionality than the original data.
The original data are not used as direct inputs to the algorithm, allowing efficient use of memory resources and storage after the initial preprocessing steps are completed.
The maximum number of eigenvectors to compute is data and application dependent, and there are almost as many different conceptions and definitions of communities in network analysis as there are problem areas.
All of the steps of the algorithm are listed in Algorithm \ref{alg:1} for the sake of clarity and reproducibility.

Once computed, the resulting Laplacian mixture models can be used as lower-dimensional or reduced representations of the original data.
This type of dimensionality reduction occurs as part of a larger machine learning system, e.g. for supervised machine learning algorithm training problems such as classification.
It can also be used for low-rank approximations in linear and multilinear (tensor) regression problems. In these cases, it is the details of the larger system that this dimensionality reduction step is embedded within that determine the amount of information loss that occurs.

This example involves single-cell expression profile data using a technique known as Drop-seq that can easily generate over 10,000 measurements per 50,000-cell tissue sample \cite{dropseq}.
Cell contents are suspended within droplets, and the mRNAs are captured on microbeads with unique DNA barcodes that allow single-cell analyses to be done in parallel without losing the association to the individual cells captured \cite{dropseqrev}.
Tissues such as the retina have specialized and differentiated cell types that have been identified as histologically distinct, and single-cell expression data potentially provides a means of matching cell types with unique gene expression features.
One of the issues in the field is its exploratory nature, requiring unsupervised machine learning approaches since there are no ``ground truth'' data or labels available (E.Z. Macosko, personal communication).

Probabilistic models can infer whether some set of low-dimensional factors can explain the various expression profiles measured.
This example represents one of the first nonhierarchical analyses of single-cell expression and potentially paves the way for useful biological insights into the structure of cellular expression patterns.

To tune the structures recognized by the first $m$ Laplacian eigenvectors, a minimum weight parameter was introduced.
This parameter adjusts the minimum similarity value.
I.e., if an item's max similarity is less than this value, it was set to it, connecting all of the items with at least one other item to prevent weakly-connected graphs.
The minimum weight parameter was set to the 99-th percentile value over all pairs for the purpose of illustrating the method.
According to  \cite{ronan2016avoiding}, this parameter should be varied across some range of interest and consensus information extracted from the ensemble.

In addition to the minimum weight parameter, there are many possible choices of normalization for the measurements prior to fitting or learning the model in an unsupervised manner.
Euclidean distances are well tested, but in multidimensional spaces they are sensitive to the choice of normalization.
To identify unpredicted biological information, ensemble clustering approaches are recommended, where multiple perturbations of the cluster analysis are made, including the parameters of clustering \cite{ronan2016avoiding}.
The denoised normalization developed for the example shown below is suggested as an additional perturbation that may be used for future ensemble clustering setting, not as a general improvement for all cases.

\begin{figure}[!ht]
\centerline{\includegraphics[width=5.0in]{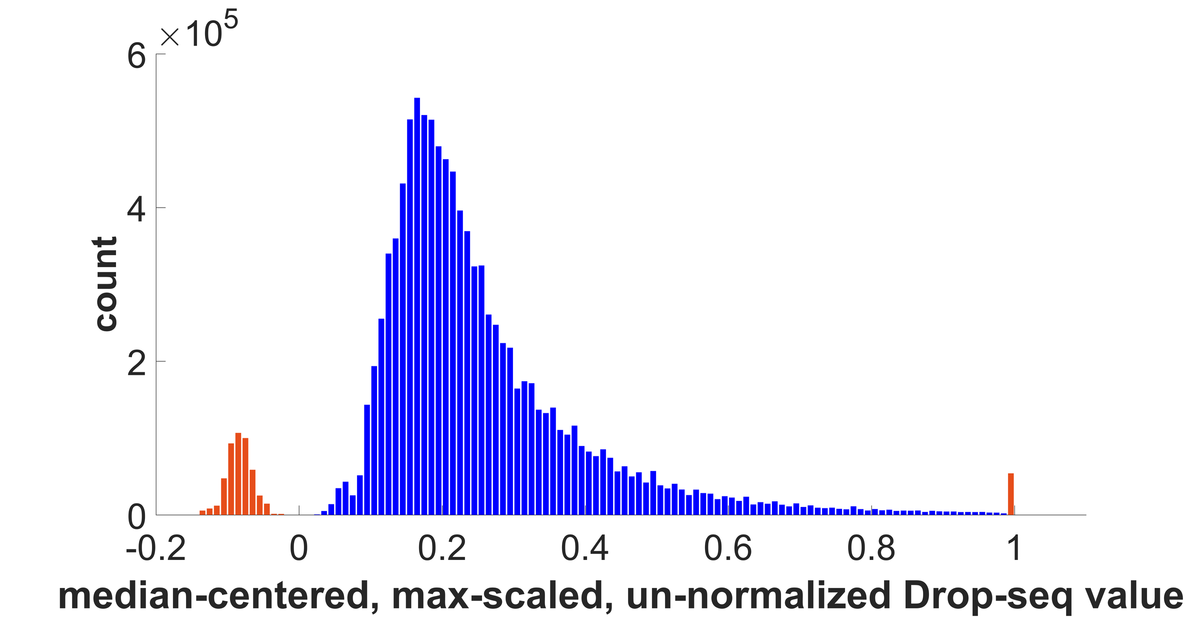}}
\caption{
{\bf Histogram of 10,900,036 nonzero Drop-seq datapoints after subtracting median and dividing by max value for each cell.}
Bins containing negative values on the left and the bin containing one on the right are colored orange to indicate outlier distributions and contain $<5$\% of the data.
Bins containing values between 0 and 0.995 are colored in blue and capture $>95$\% of the data.
Only the data from the bins shown in blue was used for subsequent data clustering steps.
}
\label{fig:2}
\end{figure}

The motivation for developing a denoised normalization is provided by the histogram Fig~\ref{fig:2}, which shows the distribution of un-denoised median-centered and max-scaled data.
After subtracting the median value, 9 cells were all zeros, i.e. these cells probably contained no useful data and were not included in the subsequent analyses to avoid spurious clusters.
(The indices of these all-zero cells in the dataset are: 4583, 6148, 13026, 15439, 17395, 24267, 26655, 28148, and 43383.)

The bulk of the data ($>95\%$) occurs in the region containing the broad peak shown in the blue colored bins, lying between the orange colored bins containing negative values and the bin containing a value of one.
Bins containing negative values, and the bin containing the value of one, in orange, do not match the shape of the portion shown in blue containing most of the data, and appear to belong to other distributions that should be analyzed separately or treated as measurement artifacts or noise.
For the purposes of the example shown here, the points indicated by orange bins in Fig~\ref{fig:2} are treated as outliers and removed as a denoising step, prior to performing any data clustering.
Since this denoising step removes less than 5\% of the total data, it is reasonable to assume it does not affect the biological relevance of the results, when the goal is to obtain a view of global patterns contained in the entire dataset as whole.
The saturated values indicated by the orange bin on the right cannot be distinguished from artifacts due to contamination or sensor malfunction.
Removing them is a conservative choice in order to focus on the dominant mode of the distribution and to make the analysis less sensitive to potential outliers.

\begin{figure}[!ht]
\centerline{\includegraphics[width=5.0in]{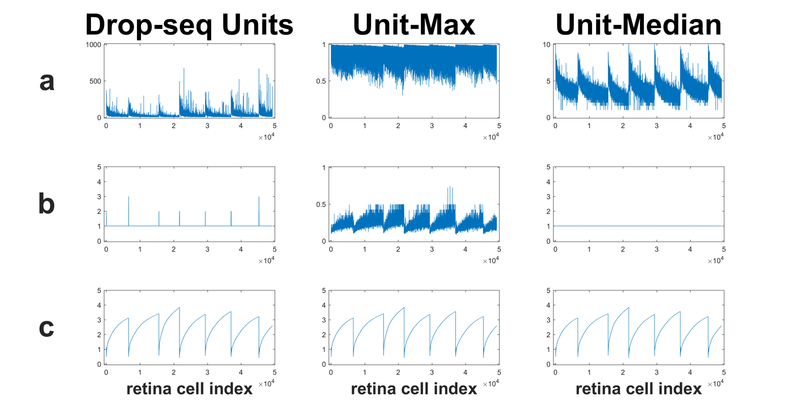}}
\caption{
{\bf Comparison of normalization in \cite{dropseq} (left column) to the denoised unit normalization used here (right column).}
The bottom row \textbf{c} shows the median of the same unnormalized columns that were input into both normalization procedures.
The middle row \textbf{b} shows the median of the normalized values for each column of data, where columns correspond to retina cell types.
Row \textbf{a} shows the maximum normalized value for each column of data.}
\label{fig:3}
\end{figure}

Another reason for using the denoised normalization developed here is because the normalization used in \cite{dropseq} contains features that may adversely affect statistical analyses that involve Euclidean distances.
Fig~\ref{fig:3} compares the denoised unit-max and unit-median normalizations for the 49299 cells (columns) and 24657 genes (rows) used here to the normalization in \cite{dropseq}.
The bottom row \textbf{c} shows the median of the unnormalized data columns, corresponding to retina cell types, for reference to line up the relevant features in the top two rows.
Rows \textbf{a} and \textbf{b} show the max and median of each cell type, respectively.
The left column shows the Drop-seq normalization and the middle and right columns respectively show the denoised unit-max and unit-median normalizations used here.

Note the alignment of the local minima of the unnormalized medians shown in row \textbf{c} with the artificial trends shown in rows \textbf{a} and \textbf{b} of the left column.
In particular, the median shows a value of 1 except at the instabilities whenever the local minima in the bottom row occur.
These spikes in the median shown in the middle row are enhanced by quantization noise, which can also adversely affect Euclidean distance based analyses.
This shows that the Drop-seq normalization may be overly and contains outliers whenever the unnormalized median reaches a local minimum.

The top row of the left column shows that the trend in the max of the normalized values creates outliers around the local minima shown in the bottom row.
This suggests that their normalization may potentially be amplifying noise to accentuate the nonlinear trend shown row \textbf{a} of the left column.
Dividing by small values is well known to increase numerical instabilities using finite-precision arithmetic.
Rows \textbf{a} and \textbf{b} of the middle and right columns show the denoised normalizations used here.
The same denoising step was used for both unit-max and unit-median normalizations.

Additive offset noise was removed by subtracting the median value, and multiplicative gain noise was removed by scaling max values to one, after correcting the offset.
After gain and offset correction, outlier noise was then identified as negative and unity values and removed.
The unit-median normalization was obtained from the denoised unit-max normalization by dividing all columns by the median of their nonzero values.
As the left and right columns if the top row of Fig~\ref{fig:3} show, this makes the distribution more similar compared to the original Drop-seq normalization.
The top row of Fig~\ref{fig:3} shows that the unit-max or unit-median normalizations (middle and right columns, respectively) are distributed within a smaller range than the Drop-seq normalization (left column).

Unlike the Drop-seq normalization, the denoised unit-max and unit-median normalizations are unquantized and cannot be interpreted as representing physical molecular copy numbers.
Denoising makes the data more appropriate for Euclidean distance based data analysis because of its sensitivity to outliers.
This makes the denoised normalizations potentially better for statistical analyses involving Euclidean distances such as $k$-means, which become more sensitive to outliers as the dimensionality of the embedding space increases.
Prior to analysis using pairwise Euclidean distance driven approaches, future studies may also convert these offset and gain corrected and denoised values to $Z$-scores, providing an additional clustering perturbation for use in ensemble approaches as described in \cite{ronan2016avoiding}.

\begin{figure}[!ht]
\centerline{\includegraphics[width=5.0in]{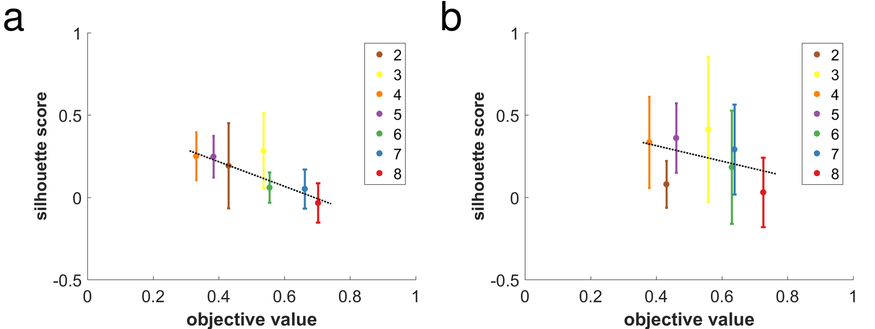}}
\caption{
{\bf 2-through-8 factor model silhouette score estimates computed by averaging over 10 sets of randomly subsampled cells (2000 cells per sample) vs. optimal objective value for (\textbf{a}) unit-max and (\textbf{b}) unit-median denoised normalizations.}
Dashed lines indicates robust linear fit computed using iteratively reweighted least squares.
The 3-factor silhouette scores (yellow) were consistently outlying above the linear trend shown by the dashed line for both normalizations, and the 7-factor solution (blue) is the highest dimensional model with positive residual.
}
\label{fig:4}
\end{figure}

Fig~\ref{fig:3} suggests that the denoised unit max normalization may provide better results when using Euclidean distances for pattern recognition because the resulting distribution is more compact.
This justifies using the unquantized denoised normalization shown on the middle and right columns for the Laplacian mixture model examples shown here, although more thorough studies are needed using ensemble clustering approaches to validate these results in the future.
In order to compare between the denoised unit-max and denoised unit-median normalizations, the residuals of a linear fit to the plot of silhouette score vs. objective value were compared.
Fig~\ref{fig:4} shows the silhouette scores for the 2-through-8 factor Laplacian mixture models generated for the (denoised) unit-max (\textbf{a}) and (denoised) unit-median (\textbf{b}) normalizations.
The dotted line shows the robust linear fit identified from the silhouette scores, indicating a negatively sloping trend suggesting that higher factor models may be overfitting.
Linear fit residuals for the unit-max normalization (\textbf{a}) are noticeably smaller in magnitude than those for the unit-median normalization suggesting it may be more stable for Euclidean-distance based network analyses.
Therefore all subsequent results shown here made use of the denoised unit-max normalization.
After computing the denoised unit-max normalization, 66 columns were found to have constant values and were removed prior to clustering, and 8653 (35\%) of denoised rows were found to be duplicates and removed.
An input data matrix of 49,233-by-16,004 values was left for computing the Laplacian that was input into the global optimization algorithm determining the corresponding Laplacian mixture model.

The $k$-nearest neighbor method with corresponding parameter $k_{NN}$ was used to set the maximum number of nonzero entries in any row or column of the resulting pairwise similarity matrix.
This ensures sparsity and prevents instabilities in the Laplacian eigenvector structure.

\begin{figure}[!ht]
\centerline{\includegraphics[width=5.0in]{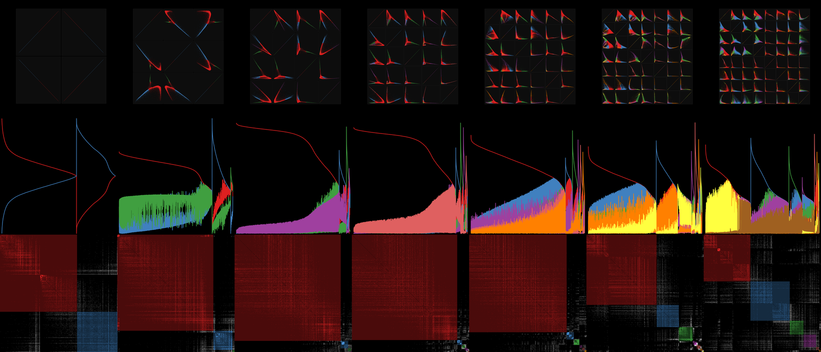}}
\caption{
{\bf Sequence of models for the unit-max normalized Drop-seq retina cell profiles published on GEO (ID GSE63472).}
Top row shows scatterplot matrices colored by thresholded cluster assignment index for 2-8 factor models.
Middle row shows corresponding factor conditional probability line plots sorted by max assignment index.
Diagonal blocks on images in the bottom row show the corresponding sorted input similarity matrix revealing hidden structure in the unlabeled data.
}
\label{fig:5}
\end{figure}

Empirically when a useful model can be computed, it is typically part of a sequence of models beyond which an acceptable (nondegenerate) solutions cannot be found.
The rigorous mathematical explanation for this has not been fully understood yet but is discussed in more detail in section \ref{sec:conclusion}.
Since no 9-factor solutions were found, the search was truncated at $m=8$.
Fig~\ref{fig:5} shows the optimized 2-through-8 factor models for the unit-max normalization.
Since the 7-factor solution showed the highest dimensionality with positive residual, it was chosen to visualize in more detail.

\begin{figure}[!ht]
\centerline{\includegraphics[width=5.0in]{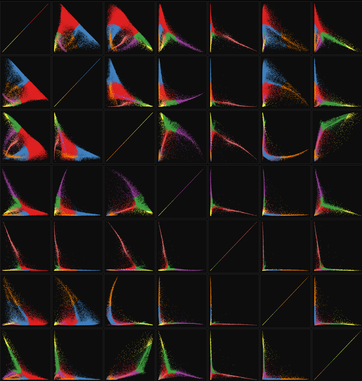}}
\caption{
{\bf Grouped scatterplot matrix of conditional probabilities for the 7-dimensional unit-max normalized Drop-seq retina cell Laplacian mixture model.}
Axes are autoscaled inside the interval [0, 1] for all panels.
Horizontal axes are aligned by columns, and vertical axes are aligned by rows.
Colors indicate max probability assignment index showing the corresponding hard clustering generated by thresholding.
Hard clustering assignment counts were for the overlapping modules detected.
}
\label{fig:6}
\end{figure}

Fig~\ref{fig:6} shows a zoom in on the scatterplot matrix for the $m=7$ solution shown in the top row of Fig~\ref{fig:5}.
Subsequent hierarchical or other hard-clustering algorithms  can be applied using these 7-dimensional conditional probabilities as feature vectors in order to generate non-overlapping clusters or communities if needed.

Colors indicate hard cluster assignments generated by thresholding appear visually reasonable, suggesting that potentially informative overlapping communities were detected.
Unlike many other community detection algorithms based on global optimization, Laplacian mixture models can identify communities with sizes that are different orders of magnitude.

\begin{figure}[!ht]
\centerline{\includegraphics[width=5.0in]{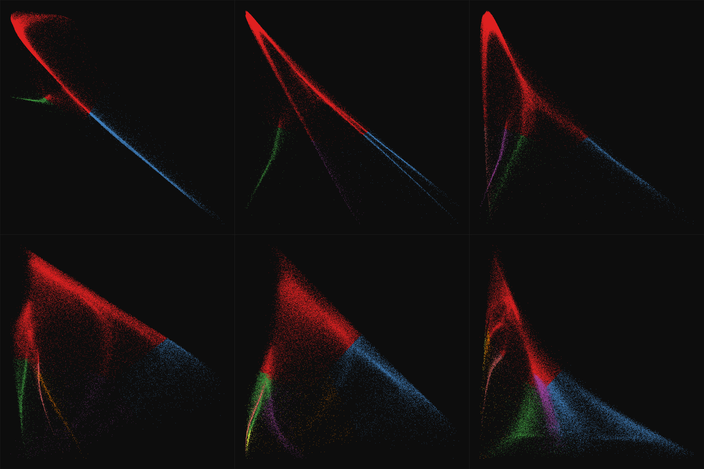}}
\caption{
{\bf Grouped scatterplots of conditional probabilities for components 1 vs. 2  from the 3-through-8 factor unit-max normalized Drop-seq retina cell Laplacian mixture models.}
Plots are in order of ascending model dimension (3 through 8) from left-to-right, top-to-bottom.
Colors indicate max probability assignment index showing the corresponding hard clustering generated by thresholding.
Axes are autoscaled inside the interval [0, 1] for all panels.
}
\label{fig:7}
\end{figure}

The grouped scatterplots for the 1st-vs-2nd conditional probabilities from the top row of Fig~\ref{fig:6} are plotted separately in Fig~\ref{fig:7}.
These highlight the differences between the models in terms of their ability to separate the data into potentially informative structures identified in the Drop-seq retinal cell profiles.
From a signal processing and graph partitioning standpoint, higher factor numbers provide more fidelity and parallelization at the expense of compression ratio.
Once such a sequence of models has been computed, standard model selection techniques from machine learning and statistics are directly applicable.
The problem of model selection will not be dealt with in detail here since the focus is to demonstrate the basic approach.
Once ground-truth data are available, these models can be selected for biological usefulness.
\subsection*{Graph/Network Analysis}
Unlike data clustering using symmetric item pairwise-similarity matrices as input, graph clustering accepts weighted adjacency matrices that are often not symmetric but can be symmetrized.
Biological networks, social patterns, and the World Wide Web are just a few of the many real world problems that can be mathematically represented and topologically studied in terms of community detection \cite{communities}.

Generally, network community detection tries to infer functional communities from their distinct structural patterns \cite{yang2015}.
Functional definitions of network communities are based on common function or role that the community members share.
Examples of functional communities are interacting proteins that form a complex, or a group of people belonging to the same social circle.

Modularity is one quantity that, when maximized, provides a measure of communities potentially having different properties such as node degree, clustering coefficient, betweenness, centrality, etc., from that of the average network.
Because modules of many different sizes often occur, the potential limits of multiresolution modularity and all other methods using global optimization have been suggested by \cite{modularity}.
In particular, methods based on global optimization have been suspected of being incapable of finding communities at many different sizes or scales simultaneously.

Although Laplacian mixture models are a global optimization type method, they are inherently multiscale since higher dimensional eigenspaces encode multiresolution dynamics from a Markov process interpretation.
In order to illustrate the promise of using Laplacian mixture models to provide useful global, nonhierarchical views of large graphs, the \emph{E. coli} genome was analyzed as a test problem.

\emph{E. coli} is a bacterial model organism in biology that has a relatively small genome but (like all genomes) one that still contains many transcribed regions with unknown functions.
Using interactome data provided by the HitPredict protein-protein interaction database \cite{hitpredict}, one large connected component was identified for analysis, containing 3257 out of 3351 total genes/proteins.

In order to prepare the weighted adjacency matrix for the analysis, the top $k_{min}$ connections were strengthened using the .99th quantile over all weights, and then a nearest-neighbor cutoff of $k_{NN}=125$ was applied.
Next, rows of the weighted adjacency matrix were normalized to sum to one, to reduce the influence of proteins with stronger connections on average.
Finally, because protein-protein interactions are mutual by nature, the matrix was symmetrized by taking the elementwise maximum between it and its transpose.

\begin{figure}[!ht]
\centerline{\includegraphics[width=5.0in]{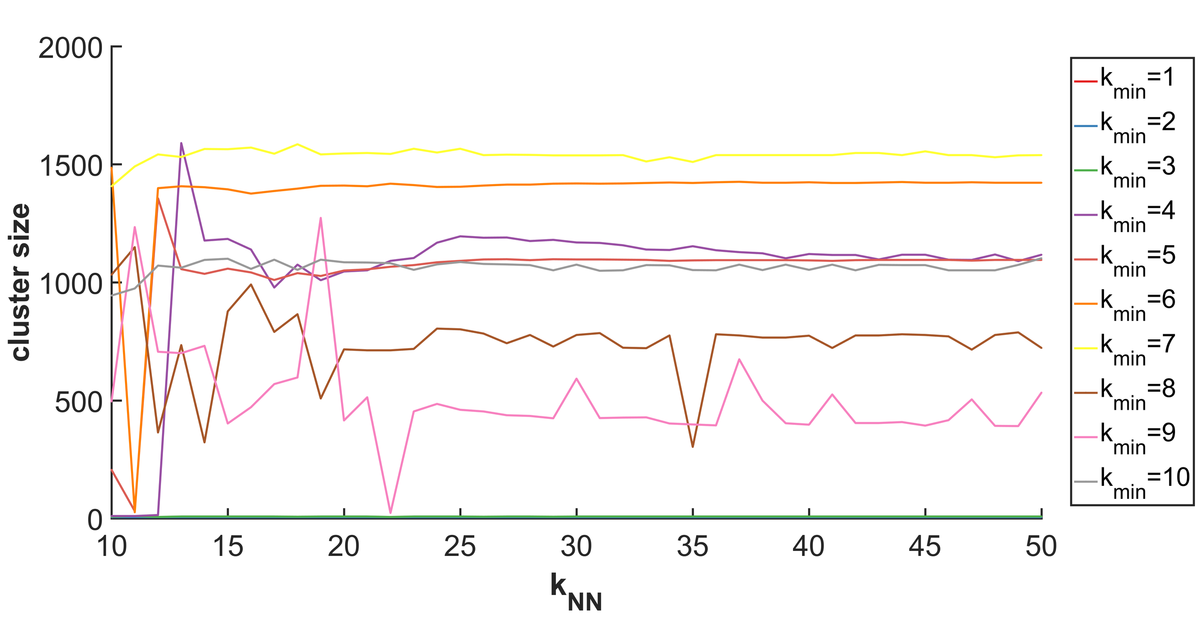}}
\caption{
{\bf Max cluster size of 5-factor models vs. $k_{NN}$ for \emph{E. coli} protein protein interaction data.}
Colors correspond to different values of $k_{min}$.
The plot shows that $k_{min}=4$ was the lowest value to show a reasonably large size for the 2nd-largest cluster after hard thresholding the conditional probabilities.
}
\label{fig:8}
\end{figure}

\begin{figure}[!ht]
\centerline{\includegraphics[width=5.0in]{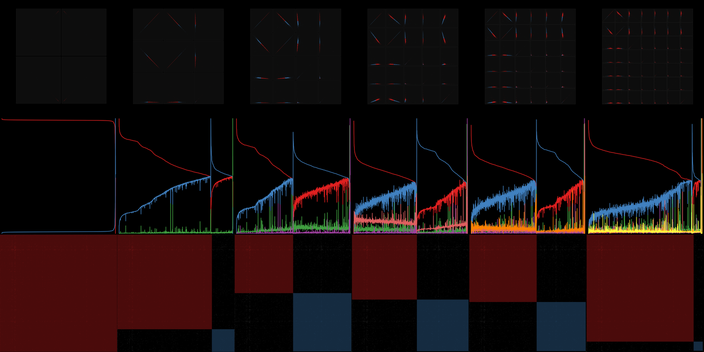}}
\caption{
{\bf \emph{E. coli} interactome analysis (3257 proteins with 20239 pairwise interactions) showing 2-through-7 factor models.}
Top panel shows grouped scatterplot matrices, middle row shows conditional probability line plots, and bottom panel shows corresponding graph adjacency matrices sorted by max conditional probability value.
}
\label{fig:9}
\end{figure}

To select appropriate values of $k_{min}$ and $k_{NN}$, a grid search was performed by computing all five-factor models over the ranges $k_{min}=3, \dots, 5$ and $k_{NN} = 1, \dots, 192$, where 192 is the maximum degree for the entire graph.
The largest cluster size after hard thresholding by maximum conditional probabilities was plotted as shown in Fig~\ref{fig:8}.
For this dataset, a sequence of 2-through-7 factor solutions was found using $k_{min}=4$ and $k_{NN}=\infty$, keeping all of the edges from the original dataset.
Fig~\ref{fig:9} provides the same three views of the sequence of models as for the Drop-seq analysis above.
Since the primary input was interaction data and no pairwise distances were computed, the silhouette scores are not available.
Standard model selection techniques can be applied at this point, but are not performed here since the focus is on demonstrating Laplacian mixture modeling approaches.
It is possible to gain some insight into model quality by examining the contributions of the individual model components to the squared loss or uncertainty score.
Table \ref{tab:1} lists the component contributions to the loss function, with lower values being more favorable.
\begin{table}[!ht]
\renewcommand{\arraystretch}{1.3}
\begin{adjustwidth}{-0.25in}{0in}
\caption{
{\bf Component losses of Laplacian mixture models for the \emph{E. coli} interactome network dataset.}
}
\centering
\begin{tabular}{l|l|l|l|l|l|l|l}
  \multicolumn{1}{c}{} & \multicolumn{7}{c}{component index}\\
  dimensionality & \multicolumn{1}{|c|}{1} & \multicolumn{1}{c|}{2}  & \multicolumn{1}{c|}{3}  & \multicolumn{1}{c|}{4}  & \multicolumn{1}{c|}{5}  & \multicolumn{1}{c|}{6}  & \multicolumn{1}{c}{7} \\
  \thickhline 
  2 & 0.912 & 0.016   &   &  &  &  &\\ \hline
  3 & 0.840 & 0.355 & 0.576 &  &  &  & \\ \hline
  4 & 0.824 & 0.510 & 0.904& 0.431&  &  &\\ \hline
  5 & 0.473& 0.519& 0.901& 0.823& 0.895&  & \\ \hline
  6 & 0.478& 0.943& 0.489& 0.824& 0.924& 0.890\\ \hline
  7 & 0.864& 0.924 & 0.907& 0.935& 0.673& 0.956& 0.361\\ \hline
\end{tabular}
\label{tab:1}
\end{adjustwidth}
\end{table}
The 6-factor model was the only one having two components less than 0.5 and all components less than 0.95.
Its hard-thresholded sizes for this model's components were 1869, 1362, 11, 7, 6, and 2.
Different orders of magnitude component sizes suggest the ability to detect multiscale communities despite belonging to the class of global optimization type methods.
The next steps in developing this method could include delving into the possible biological significance of the patterns identified by Laplacian mixture models in an unsupervised way by analyzing the \emph{Homo sapiens} interactome.
\subsection*{Density Estimation}\label{sec:density}
Nonparametric mixture density separation is a challenging problems in statistics.
In the context of mixture density function separation or unmixing problems, Laplacian mixture models fall into the class of partial differential equation (PDE)-based methods.
The connections of Laplacian mixture models to PDE ``coarse-graining'' methods are described in more detail in the appendix.

These differential equations allow Laplacian eigenspaces to be defined from input mixture density estimates as described below.
The resulting Laplacian mixture models define globally optimized mixture component estimates directly from the spectral information contained in the discretized PDE.

This synthetic mixture density separation example allows unambiguous evaluation of the performances of different tuning parameter choices.
For the first example, a randomized three-component mixture $f(x) \propto f_1(x) + f_2(x) + f_3(x)$ was constructed, consisting of one component made from a mixture of three radial basis functions with randomized covariance transforms: Gaussians, Laplace distributions, and hyperbolic-secant functions.
Since these components do not share any common parametrizations, using any one class of distribution to compute the unmixing will result in errors.
This simple yet nontrivial example provides a test of the accuracy of the nonparametric Laplacian mixture modeling approach for nonparametric density function unmixing.
that include additional smoothness or regularity assumptions.

\begin{figure}[!ht]
 \centerline{\includegraphics[width=4.5in]{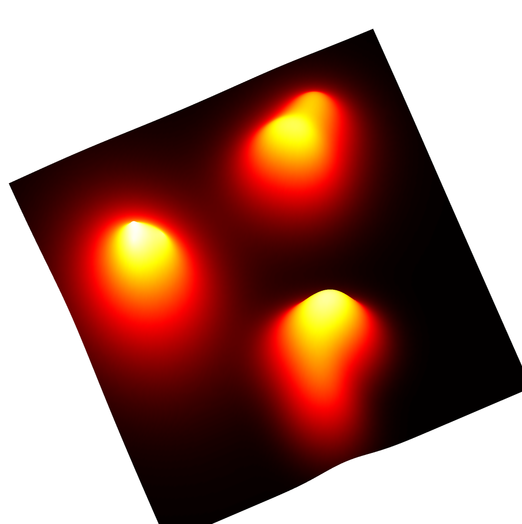}}
 \caption{
 {\bf Gaussian/Laplace/hyperbolic-secant mixture density function surface plot colored by probability density.}
 Each of the three separable components were constructed by
 adding randomly generated anisotropic radial functions with either Gaussian, Laplacian, or hyperbolic-secant radial profiles. Finally, these randomized components were
 superimposed to generate the final mixture distribution shown here.}
 \label{fig:10}
 \end{figure}

Discretized density function values are taken as input for Laplacian mixture model computation via direct approximation of a class of heat equations known as Smoluchowski equations in physics as described in
\ref{sec:dapprox}. As explained in \cite{banush},  the input mixture density estimate $f(x)$ is uniformly sampled on a discrete grid or lattice of points using neighbor indices $I_i, i=1, \dots, N.$
where $\beta \in (0, \infty)$ acts as a scaling parameter with the interpretation of inverse absolute temperature in statistical physics.
Non-boundary off-diagonal Laplacian matrix values are then set according to
\begin{equation}
q_{ij} = \left\{
{\setlength{\extrarowheight}{6pt}
    \begin{array}{cc}
    \exp{\left[ \frac{\beta}{2}\log \frac{f(x_i)}{f(x_j)}\right]}, & i < j \in I_i \\
    \exp{\left[ \frac{\beta}{2} \log \frac{f(x_j)}{f(x_i)} \right] }, & i > j \in I_i \\
    0, & \mathrm{otherwise}.
    \end{array}
}
\right.
\end{equation} 
Fig~\ref{fig:10} shows an image of $f(x)$ evaluated at $40,000$ Cartesian gridpoints $\{x_i: x_i \in [-10, 10]^2\}_{i = 1}^{200}$ in two dimensions with colors
indicating the value of $f(x)$ at each point.
Algorithm performance can be improved by tuning the $\beta$ parameter. The topological structure of macrostate splitting as $\beta$ increases from sufficiently small nonzero value towards $\infty$ has been useful for solving challenging global optimization problems \cite{pard94}.
Such homotopy-related aspects of macrostate theory are not explored here for the sake of brevity.

\begin{figure}[!ht]
 \centerline{\includegraphics[width=4.5in]{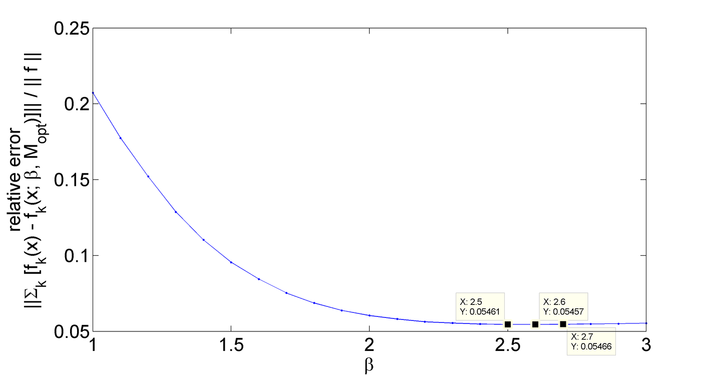}}
 \caption{
 {\bf Relative error of the $m=3$ Laplacian mixture model for the Gaussian/Laplace/hyperbolic-secant mixture test problem vs. $\beta$.}
 Minimum value of $\beta=2.6$ indicated by flanking by datatips.
 }
 \label{fig:11}
 \end{figure}

\begin{figure}[!ht]
 \centerline{\includegraphics[width=4.0in]{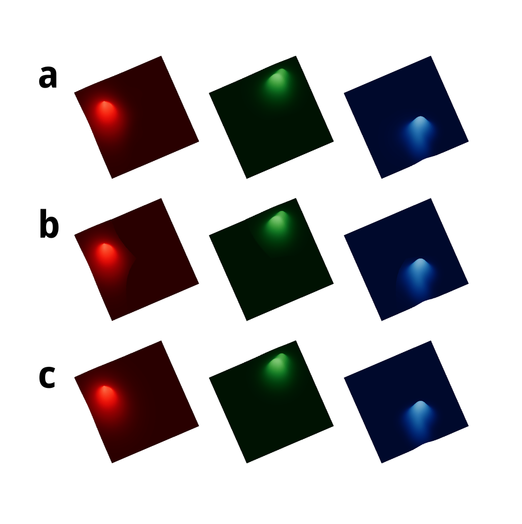}}
 \caption{
 {\bf Optimally-scaled $\beta=2.6$ Laplacian mixture model components for the Laplace (red)/hyperbolic-secant (green)/Gaussian (blue) 2-D test problem.}
 Row a: unthresholded Laplacian mixture model components,
 Row b: hard-thresholded components,
 Row c: original (unmixed) components. Column 2 of Table \ref{tab:1} lists the corresponding mean squared errors.}
 \label{fig:12}
 \end{figure}

Since the true solution was known for this test problem, the relative error vs. $\beta$ was optimized with a one-dimensional grid search as shown in Fig~\ref{fig:11}.
Fig~\ref{fig:12} shows the optimized Laplacian mixture model for this test problem.
Rows a and b of Fig~\ref{fig:12} appear acceptable with no visible mixing across components.
Column 2 of Table \ref{tab:2} shows the relative errors for the probabilistic/unthresholded and hard-thresholded optimal $\beta=2.6$ solutions.
Notice that hard-thresholding these components along their decision boundaries increases their error value compared to the original soft/probabilistic/fuzzy model.
\begin{table}[!ht]
\begin{adjustwidth}{-0.25in}{0in}
\caption{
{\bf Relative errors of Laplacian mixture models for the Gaussian/Laplace/hyperbolic-secant mixture density function separation/unmixing test problem.}
}
\centering
\begin{tabular}{l|l|l}
  \multicolumn{1}{c}{} & \multicolumn{2}{c}{$\beta$}\\
  & \multicolumn{1}{|c|}{1} & \multicolumn{1}{c}{2.6} \\
  \thickhline
  no threshold   & 0.2078 & 0.0541 \\ \hline
  hard threshold & 0.1803 & 0.0718 \\ \hline
\end{tabular}
\label{tab:2}
\end{adjustwidth}
\end{table}
\section*{Conclusion}\label{sec:conclusion}
Laplacian mixture models are a new tool for probabilistic/fuzzy spectral clustering and graph/network analysis because they nonhierarchically identify large separable regions and their interconnections.
This provides high level soft partitions or dissections of big/massive data in their entirety without using any iterative/localized seed points.
Their versatility and flexibility come at the cost of their computational and model selection challenges.
Since their original formulation in \cite{korenblum}, many implementation challenges have been overcome and their connections to other Laplacian eigenspace methods have been developed, leading to the current reformulation including new loss functions and notation.

Many possible models can be computed and model selection is an important consideration in order to select the subset of models that are most appropriate for a given application.
In some applications such as compression or denoising, minimizing the number of components, partitions, clusters, or factors is more important than perfect reproduction of the original signal.
Other applications such as recovery of a reference signal might benefit from choosing the largest number of factors that are computationally feasible.
The optimal choice model may also be constrained by relative to available resources for a given application.
Model selection is application dependent and the examples presented here may not provide the best results for every problem but may be useful as a guide for future studies.

Laplacian eigenspaces impose constraints on the space of possible models that can be defined, providing a form of spectral regularization.
In the context of data clustering, eigenspace structures are determined by the choice of distance or similarity measure and the choice of parameters used for this measure.
For graph/network analysis, the process of computing a weighted adjacency matrix can be adjusted to fine-tune the corresponding Laplacian eigenspace structure.
Applications involving unmixing mixture distribution function estimates can tune a parameter in the corresponding partial differential equation used to define the Laplacian.

Another potential application not demonstrated in the examples section includes more accurate information retrieval, search, and recommender systems.
The PageRank or Google algorithm and its personalized or localized variants have become standard methods in these application areas \cite{gleichpage}.
Originally, PageRank was used for ranking search results according to overall graph centrality score as given by the Perron-Frobenius eigenvector.
The original PageRank algorithm provided a scalable and practical method for large graph datasets such as the World Wide Web, but nodes with similar centrality scores might not belong to the same location in the graph.
Later, personalized and localized variants of the original PageRank algorithm were developed to address this issue, but introduce bias from the choice of seed nodes or locations and lose the global breadth of the original PageRank method.
Laplacian mixture models may provide a regional PageRank variant, refining the original PageRank centrality scores according to the regions of the graph encoded by the mixture components, a possibility left for future work.

Regardless of the application or type of input data structure, optimizing the structure of Laplacian eigenspaces can be challenging to do manually and may be difficult to fully automate.
While Laplacian mixture models are nonparametric in terms of the data-dependent functions their eigenspaces support, in practice there are tuning parameters involved for processing raw input data.
Several strategies for semi-automatically optimizing the Laplacian eigenspace structure were presented here, and more work in this area will be done in the future.

Laplacian eigenspaces explain all of the nonequilibrium dynamics defined by the Markov chain generated by the corresponding Laplacian, and hence have dynamic interpretations.
They also have physical interpretations, where the PF eigenvector represents a fixed point of the differential equation from a dynamic systems perspective.
The physical and dynamic systems interpretations of Laplacian eigenpaces complement their statistical and algebraic interpretations, revealing new connections between ideas from previously separate fields.
Laplacian mixture models are an example of how combining ideas from physics, and statistics provides valuable new data analysis algorithms, where many connections remain to be found.

During the global optimization step, many more models are computed and then only a subset (often one) is selected from these samples.
Different loss functions can affect which solution(s) are accepted from the output of the global optimization algorithm, and the quadratic loss function used here can be modified freely depending on the application details.
The squared loss function has been empirically verified to provide reasonably good models for a wide variety of data inputs, and similarly validating other loss functions would be of value.
It also generates a linearly-constrained concave quadratic global optimization problem which has been well-studied in the literature and can be approximately solved with high accuracy for certain convex hull geometries.

The provably optimal recovery results for noise-free interpolating cluster graphs provides an absolute mathematical reference for the algorithm's performance in ideal settings.
Performance comparison to other community detection and spectral graph partitioning algorithms are needed, with a wider variety of test datasets.
Subsequent studies will focus on thorough evaluation of various performance metrics to establish the pros and cons of this algorithm compared to other algorithms empirically.

Future studies will focus on empirically developing a better understanding of how the geometry of the convex hull formed by the linear inequality constraints relates to the statistical qualities of the resulting models.
The next step will be a detailed experimental analysis of the pros and cons of Laplacian mixture models for item data, graph/network data, and density/distribution function data using a thorough set of benchmarks.
Perhaps one day this story will circle back to its mathematical origins in Perron-Frobenius theory and create a picture connecting mathematics, physics, statistics, and machine learning.
An intuitive formal theory to guide the development has been presented here, but the rigorous theory is incomplete.
\section*{Acknowledgments}
I sincerely thank David Shalloway, Walter Murray, and Michael Saunders for their instruction and advice, Panos Pardalos for suggesting the use of the modified Frank-Wolfe heuristic, and to David Gleich and Felix Kwok for helpful discussions.
I also thank the reviewer Jingrui He, the anonymous reviewers, and the journal editors for useful suggestions and improvements for revisions and corrections.
\section*{Appendix}
\subsection*{Macrostates}
The original formulation of Laplacian mixture models comes from the definition of macrostates of classical stochastic systems, in equation (24) of \cite[page 9990]{shall96}, restated as
\begin{equation}
\psi_0(x) = \sum_{k = 1}^m p_k \Phi_k(x).\label{eq:mxpn}
\end{equation}
where the coordinate vector $R$ from the original statement has been replaced by $x$ and $k$ replaces the index $\alpha$, to match the notation used here.
The $\psi_i(x) \equiv \phi_i(x) / \sqrt{\phi_0(x)}, i=0, \dots, N$ are equivalent to the Laplacian eigenfunctions.
The mixture components $\Phi_k(x)$ were originally defined by equation (26) of \cite{shall96}, after substituting $\alpha \leftarrow k$ as in \eqref{eq:mxpn}, as
\begin{equation}
\Phi_k(x; M) = \sum_{i = 0}^{m - 1} M_{k i} \psi_i(x).
\end{equation}

Multiplying both sides by $\psi_0(x)$ to match the definition
\begin{equation}
  \hat f_k(x) \equiv \psi_0(x) \Phi_k(x) \label{eq:fhatdef}
\end{equation}
recovers the form of \eqref{eq:fmm}, a finite mixture model, in terms of the Laplacian eigenspace $\{\phi(x)_i\}_{i = 0}^m$.

Macrostates were originally created to rigorously define both the concept of metastability and also the physical mixture component distributions based on slow and fast time scales in the relaxation dynamics of nonequilibrium distributions to stationary states \cite{shall96}.
Mixture models are used for linearly separating these stationary or Boltzmann distributions in systems with nonconvex potential energy landscapes where minima on multiple size scales occur, e.g. high-dimensional overdamped drift-diffusions, such as macromolecules in solution.
Proteins folding, unfolding, and aggregating in aqueous solution are one type of biological macromolecule that can be described in terms of overdamped drift-diffusions \cite{shall96}.

Transitions between states belonging to different components of a mixture occur on relatively slow timescales in such systems, making them appear as the distinct discrete states of a finite-state continuous time Markov process when measured over appropriate timescales.
Such systems are called \emph{metastable} \cite{risken, shall96}.

In the macrostate definition, the variable $x$ is continuous and the Markov process is a continuous-state, continuous-time type known as a drift-diffusion in physics.
Mathematically, drift-diffusions are described as a type of continuous-time Markov process analogous to CTMCs, and samples or stochastic realizations of drift-diffusions are described by systems of stochastic differential equations known as \emph{Langevin equations} \cite{risken}.
For the purposes of using Laplacian mixture models, it is sufficient to know that the eigenvectors of the Laplacian are analogous to the eigenfunctions of Smoluchowski or heat/diffusion operators \cite{smola2003kernels,belkin2003laplacian,kondor2002diffusion}.
\subsection*{Smoluchowski Equations}
The Laplacian mixture component estimates $\hat f_k(x)$ are defined as expansions of $\{\phi_i(x)\}_{i = 0}^{m - 1}$ the Laplacian eigenvectors.
For continuous mixture density functions, there is a continuous-space partial differential equation (PDE) analog of Laplacian matrices known as drift-diffusion or Smoluchowski operators \cite{risken}, a type of heat equation.


Smoluchowski equations have the form
\begin{equation}
\frac{\partial P(x, t; \beta)}{\partial t} = D \nabla \cdot e^{-\beta V(x)} \nabla e^{\beta V(x)} P(x, t; \beta)\label{eq:smol} 
\end{equation}
and belong to a class of reversible continuous-time, continuous-state Markov processes used to describe multiscale and multidimensional physical systems that can exhibit metastability \cite{risken}.


The potential energy function $V(x)$ determines the deterministic drift forces acting on ensemble members i.e. sample paths or realizations of this stochastic process and can often be viewed as a fixed parameter that defines the system structure.
The drift forces bias the temporal evolution of initial distributions $P(x, 0)$ to flow towards regions of lower energy as $t$ increases compared to free diffusion (Brownian motion).

Technically there is a different Smoluchowski-type heat equation for each distinct choice of $V(x)$.
Hence the plural form should be used generally although this is often overlooked in the literature.
\subsection*{Smoluchowski Operators}
The elliptic differential operator
\begin{equation}
L_0 \equiv D \nabla \cdot e^{-\beta V(x)} \nabla e^{\beta V(x)}\label{eq:L0} 
\end{equation}
has $e^{-\beta V(x)}$ as an eigenfunction with eigenvalue zero, also called the stationary state or unnormalized Boltzmann distribution.
It is easy to evaluate $L_0 \left[ e^{-\beta V(x)}\right]$ directly to verify that it equals zero, satisfying the eigenvalue equation.

$L_0$ is a normal operator and therefore a similarity transform $S^{-1} L_0 S$ to a self adjoint form $L$ exists \cite{risken}.
$S$ has the simple form of a multiplication operator with kernel $e^{-\frac{1}{2}\beta V(x)}$, giving
\begin{equation}
L \equiv D e^{ \frac{1}{2}\beta V(x)} \nabla \cdot e^{-\beta V(x)} \nabla e^{ \frac{1}{2}\beta V(x)}\label{eq:L} = \left[\sqrt{D} \left(\nabla - \frac{\beta}{2} \nabla V \right) \right] \cdot \left[ \sqrt{D} \left( \nabla + \frac{\beta}{2} \nabla V \right) \right]. 
\end{equation}

The stationary state of $L$ from equation \eqref{eq:L} is denoted
\begin{equation}
\psi_0(x) \equiv e^{{ -\frac{\beta}{2} V(x)}} \label{eq:psi0}
\end{equation}
and is used along with other eigenfunctions of $L$ in the separation/unmixing of the Boltzmann distribution into $m$ macrostates, analogous to the Perron-Frobenius eigenvector in Laplacian mixture models.
Adapting the notation slightly from \cite{shall96}, the eigenfunctions of $L$ are denoted $\{\psi_i\}_{i = 0}^\infty$ and the eigenvalues are denoted as $\{\lambda_i\}_{i = 0}^\infty$.
\subsection*{Discrete Approximation}\label{sec:dapprox}
For sufficiently low dimensional nonnegative functions evaluated on evenly-spaced grids, the discrete approximation of \eqref{eq:smol} can be used via a nearest-neighbor Laplacian approximation to construct a sparse approximation of $L$ in \eqref{eq:L} with reflecting boundary conditions as described in \cite{banush}.
The discrete approximation approach is useful for applications where the mixture function $f(x)$ can be evaluated on a grid such as density estimates generated by histograms or kernel density estimation.
This was the method used for the numerical example described in Section \ref{sec:density}.

Discrete approximations can also be applied to nonnegative signals such as spectral density estimates and 2 and 3 dimensional images sampled on evenly-spaced nodes after preprocessing to remove random noise.
Since discrete approximations of Smoluchowski or heat equations are microscopically reversible continuous-time Markov chains (CTMCs), macrostate models can also be constructed by embedding input data into Markov chains.

Just like in the continuous case for \eqref{eq:L0}, discrete transition rate matrices for time reversible processes are similar to symmetric matrices.
Similarly their eigenvalues and eigenvectors are real and the eigenvectors corresponding to distinct eigenvalues are orthogonal.
\subsection*{Previous Formulations}
The \emph{macrostate data clustering} algorithm is an earlier formulation developed in \cite{korenblum, white}.
Detailed comparisons between the two formulations are not included here because the previous methods required customized algorithm implementations that limited their practicality. 
These earlier papers did not explicitly mention that macrostates are a type of finite mixture model of the form \eqref{eq:fmm}, nor did they mention the Bayes classifier posterior form and probabilistic interpretations of \eqref{eq:pclass}.

Previous formulations used a different objective function and a different global optimization solver.
The objective function was a logarithm of the geometric mean uncertainty which led to a nonlinear optimization problem that was not as well-studied as quadratic programs.

Here, the original the objective function from \cite{shall96} is used, providing a more standard concave quadratic programming (QP) problem that can be more easily solved.
Linearly constrained concave QPs can be solved using established heuristics such as modified Frank-Wolfe type procedures \cite{pard93}.

Another difference is that in \cite{korenblum, white} only unbounded inverse quadratic similarity measures and soft Gaussian thresholds that do not directly control sparsity were tested.
Here, other choices of similarity/distance measures are tested and the use of hard thresholding is examined to directly control sparsity of the resulting Laplacian matrix used as the primary input into the algorithm.

The previous formulation defined in \cite{korenblum} did not mention the applications to density function unmixing/separation via \eqref{eq:mmm} and the connection to discrete approximations of Smoluchowski equations as described in section \ref{sec:dapprox}.
\subsection*{Data Spectroscopy}
In some examples tested (data not shown), hard-thresholded Laplacian mixture model results were found to agree perfectly with the output of another algorithm, called Data Spectroscopy \cite{shi09}.
Data spectroscopy does not provide a full probabilistic model including the soft/fuzzy cluster assignment probabilities and involves kernel-specific heuristics for choosing the appropriate cluster number.
But, at the level of the hard/crisp cluster labels, Data Spectroscopy is an algorithm can provide accurate estimates of hard thresholded Laplacian mixture model solutions when the same Laplacian matrices are used.

This was an unexpected outcome worthy of better understanding and more study.
The Data Spectroscopy software (DaSpec) was obtained online from the original author's website.

The mathematical arguments used to prove the accuracy of data spectroscopy in \cite{shi09} and other kernelized spectral clustering methods described more recently in \cite{schi15} may yield better understanding of the assumptions used in Laplacian mixture models as well.
Likewise, the physical interpretations of macrostates in terms of drift-diffusions and the relationship of the kernel scaling parameter to the temperature or energy of the Brownian motion of generating stochastic processes may provide additional insight into the accuracy of the approximations used by data spectroscopy.
It may be possible to hybridize Laplacian mixture models and data spectroscopic methods so that they can be used consistently on different analyses within the same project.
For example, data spectroscopy could be used during the distance/similarity/kernel function learning step.


\begin{thebibliography}{10}

\bibitem{ev96}
Everitt BS.
\newblock An introduction to finite mixture distributions.
\newblock Statistical Methods in Medical Research. 1996;5:107--127.

\bibitem{bishop}
Bishop CM.
\newblock Pattern recognition and machine learning.
\newblock Springer; 2006.

\bibitem{jordan06}
McAuliffe JD, Blei DM, Jordan MI.
\newblock Nonparametric empirical Bayes for the Dirichlet process mixture
  model.
\newblock Stat Comput. 2006;16:5--14.

\bibitem{azran}
Azran A, Ghahramani Z.
\newblock Spectral methods for automatic multiscale data clustering.
\newblock In: Computer Vision and Pattern Recognition, 2006 IEEE Computer
  Society Conference on. vol.~1. IEEE; 2006. p. 190--197.

\bibitem{njw}
Ng AY, Jordan MI, Weiss Y.
\newblock On spectral clustering: Analysis and an algorithm.
\newblock In: Advances in neural information processing systems; 2002. p.
  849--856.

\bibitem{nie2010efficient}
Nie F, Huang H, Cai X, Ding CH.
\newblock Efficient and robust feature selection via joint L2, 1-norms
  minimization.
\newblock In: Advances in neural information processing systems; 2010. p.
  1813--1821.

\bibitem{gould14}
Gould S, Zhao J, He X, Zhang Y.
\newblock Superpixel graph label transfer with learned distance metric.
\newblock In: European Conference on Computer Vision. Springer; 2014. p.
  632--647.

\bibitem{dimitriev1945characteristic}
Dimitriev N, Dynkin E.
\newblock On the characteristic numbers of a stochastic matrix.
\newblock In: Dokl. Akad. Nauk SSSR. vol.~49; 1945. p. 159--162.

\bibitem{chung1998coverings}
Chung F, Yau ST.
\newblock Coverings, heat kernels and spanning trees.
\newblock Journal of Combinatorics. 1998;6:163--184.

\bibitem{smola2003kernels}
Smola AJ, Kondor R.
\newblock Kernels and regularization on graphs.
\newblock In: Learning theory and kernel machines. Springer; 2003. p. 144--158.

\bibitem{belkin2003laplacian}
Belkin M, Niyogi P.
\newblock Laplacian eigenmaps for dimensionality reduction and data
  representation.
\newblock Neural computation. 2003;15(6):1373--1396.

\bibitem{kondor2002diffusion}
Kondor RI, Lafferty JD.
\newblock Diffusion Kernels on Graphs and Other Discrete Input Spaces.
\newblock In: Proceedings of the Nineteenth International Conference on Machine
  Learning. ICML '02. San Francisco, CA, USA: Morgan Kaufmann Publishers Inc.;
  2002. p. 315--322.

\bibitem{bett2015diffusion}
Bett DK, Mondal AM.
\newblock Diffusion kernel to identify missing PPIs in protein network
  biomarker.
\newblock In: Bioinformatics and Biomedicine (BIBM), 2015 IEEE International
  Conference on. IEEE; 2015. p. 1614--1619.

\bibitem{pearson}
Pearson K.
\newblock Contributions to the mathematical theory of evolution.
\newblock Philosophical Transactions of the Royal Society of London A.
  1894;185:71--110.

\bibitem{fasshauer2007meshfree}
Fasshauer G.
\newblock Meshfree approximation methods with MATLAB.
\newblock Singapore Hackensack, N.J: World Scientific; 2007.

\bibitem{nie2010flexible}
Nie F, Xu D, Tsang IWH, Zhang C.
\newblock Flexible manifold embedding: A framework for semi-supervised and
  unsupervised dimension reduction.
\newblock IEEE Transactions on Image Processing. 2010;19(7):1921--1932.

\bibitem{7976386}
Wang R, Nie F, Hong R, Chang X, Yang X, Yu W.
\newblock Fast and Orthogonal Locality Preserving Projections for
  Dimensionality Reduction.
\newblock IEEE Transactions on Image Processing. 2017;26(10):5019--5030.

\bibitem{kramer1959symmetrizable}
Kramer H.
\newblock Symmetrizable Markov matrices.
\newblock The Annals of Mathematical Statistics. 1959;30(1):149--153.

\bibitem{parter1962symmetrization}
Parter SV, Youngs J.
\newblock The symmetrization of matrices by diagonal matrices.
\newblock Journal of Mathematical Analysis and Applications.
  1962;4(1):102--110.

\bibitem{chung1997spectral}
Chung F.
\newblock Spectral graph theory. vol.~92 of CBMS Regional Conference Series in
  Mathematics.
\newblock Published for the Conference Board of the Mathematical Sciences,
  Washington, DC; by the American Mathematical Society, Providence, RI; 1997.

\bibitem{shamir2004cluster}
Shamir R, Sharan R, Tsur D.
\newblock Cluster graph modification problems.
\newblock Discrete Applied Mathematics. 2004;144(1-2):173--182.

\bibitem{maas1987perturbation}
Maas C.
\newblock Perturbation results for the adjacency spectrum of a graph.
\newblock Zeitschrift Fur Angewandte Mathematik Und Mechanik.
  1987;67(5):T428--T430.

\bibitem{rowlinson1990more}
Rowlinson P.
\newblock More on graph perturbations.
\newblock Bulletin of the London Mathematical Society. 1990;22(3):209--216.

\bibitem{guo2007laplacian}
Guo JM.
\newblock The Laplacian spectral radius of a graph under perturbation.
\newblock Computers \& Mathematics with Applications. 2007;54(5):709--720.

\bibitem{fiedler1973algebraic}
Fiedler M.
\newblock Algebraic connectivity of graphs.
\newblock Czechoslovak mathematical journal. 1973;23(2):298--305.

\bibitem{pard93}
Pardalos PM, Guisewite G.
\newblock Parallel computing in nonconvex programming.
\newblock Annals of Operations Research. 1993;43(2):87--107.

\bibitem{von2007tutorial}
Von~Luxburg U.
\newblock A tutorial on spectral clustering.
\newblock Statistics and computing. 2007;17(4):395--416.

\bibitem{dropseq}
Macosko EZ, Basu A, Satija R, Nemesh J, Shekhar K, Goldman Melissa, et~al.
\newblock Highly parallel genome-wide expression profiling of individual cells
  using nanoliter droplets.
\newblock Cell. 2015;161(5):1202--1214.

\bibitem{dropseqrev}
Heath JR.
\newblock Nanotechnologies for biomedical science and translational medicine.
\newblock Proceedings of the National Academy of Sciences.
  2015;112(47):14436--14443.

\bibitem{ronan2016avoiding}
Ronan T, Qi Z, Naegle KM.
\newblock Avoiding common pitfalls when clustering biological data.
\newblock Science Signaling. 2016;9(432):re6--re6.

\bibitem{communities}
Porter MA, Onnela JP, Mucha PJ.
\newblock Communities in networks.
\newblock Notices of the AMS. 2009;56(9):1082--1097, 1164--1166.

\bibitem{yang2015}
Yang J, Leskovec J.
\newblock Defining and evaluating network communities based on ground-truth.
\newblock Knowledge and Information Systems. 2015;42(1):181--213.

\bibitem{modularity}
Lancichinetti A, Fortunato S.
\newblock Limits of modularity maximization in community detection.
\newblock Physical review E. 2011;84(6):066122.

\bibitem{hitpredict}
Patil A, Nakai K, Nakamura H.
\newblock HitPredict: a database of quality assessed protein--protein
  interactions in nine species.
\newblock Nucleic acids research. 2011;39(suppl\_1):D744--D749.

\bibitem{banush}
Banushkina P, Schenk O, Meuwly M.
\newblock Efficiency considerations in solving smoluchowski equations for rough
  potentials.
\newblock In: Computational Life Sciences. Springer; 2005. p. 208--216.

\bibitem{pard94}
Pardalos PM, Shalloway D, Xue G.
\newblock Optimization methods for computing global minima of nonconvex
  potential energy functions.
\newblock Journal of Global Optimization. 1994;4(2):117--133.

\bibitem{korenblum}
Korenblum D, Shalloway D.
\newblock Macrostate data clustering.
\newblock Physical Review E. 2003;67(5):056704.

\bibitem{gleichpage}
Gleich DF.
\newblock PageRank beyond the Web.
\newblock SIAM Review. 2015;57(3):321--363.

\bibitem{shall96}
Shalloway D.
\newblock Macrostates of classical stochastic systems.
\newblock The Journal of chemical physics. 1996;105(22):9986--10007.

\bibitem{risken}
Risken H.
\newblock The Fokker Planck Equation.
\newblock In: The Fokker-Planck Equation. Springer; 1996. p. 63--95.

\bibitem{white}
White BS, Shalloway D.
\newblock Efficient uncertainty minimization for fuzzy spectral clustering.
\newblock Physical Review E. 2009;80(5):056705.

\bibitem{shi09}
Shi T, Belkin M, Yu B.
\newblock Data spectroscopy: Eigenspaces of convolution operators and
  clustering.
\newblock The Annals of Statistics. 2009;37:3960--3984.

\bibitem{schi15}
Schiebinger G, Wainwright MJ, Yu B.
\newblock The geometry of kernelized spectral clustering.
\newblock The Annals of Statistics. 2015;43(2):819--846.

\end{thebibliography}

\end{document}